\documentclass[10pt,onecolumn]{article}

\usepackage[utf8]{inputenc} 
\usepackage{longtable}
\usepackage{bm}
\usepackage{amsmath}
\usepackage{amssymb}
\usepackage{amsthm}
\usepackage{lmodern} 
\usepackage{fix-cm}  

\usepackage[dvips]{graphicx}
\usepackage{algorithm}
\usepackage{algorithmic}
\renewcommand{\algorithmicrequire}{\textbf{Input:}}
\renewcommand{\algorithmicensure}{\textbf{Output:}}
\theoremstyle{definition}
\newtheorem{lemma}{Lemma}
\theoremstyle{definition}
\newtheorem{theorem}{Theorem}
\theoremstyle{definition}

\theoremstyle{definition}

\theoremstyle{definition}

\theoremstyle{theorem}

\usepackage{color}
\definecolor{red}{rgb}{1,0,0}
\definecolor{blue}{rgb}{0,0,1}

\newcommand{\vp}{{\bm{p}}}

\newcommand{\cI}{{\mathcal{I}}}

\newcommand{\cJ}{{\mathcal{J}}}

\newcommand{\bN}{{\mathbb{N}}}

\newcommand{\bR}{{\mathbb{R}}}

\newcommand{\cS}{{\mathcal{S}}}

\newcommand{\argmax}{\operatornamewithlimits{argmax}}

\begin{document}

\begin{center}
{\Large
New Descriptor for Glomerulus Detection \\
in Kidney Microscopy Image}
\\
\vspace{1cm}
{\large Tsuyoshi Kato${}^{\dagger,\ddagger,*}$, Raissa Relator${}^{\dagger}$, 
Hayliang Ngouv${}^{\dagger}$, \\
Yoshihiro Hirohashi${}^{\dagger}$, 
Tetsuhiro Kakimoto${}^{\diamond}$, Kinya Okada${}^{\diamond}$. 
}
\\
\vspace{1cm}
\begin{tabular}{lp{0.7\textwidth}}
${}^\dagger$ & 
Faculty of Science and Engineering, Gunma University, 
Kiryu-shi, Gunma, 326--0338, Japan.  
\\
${}^\ddagger$ &
Center for Informational Biology, Ochanomizu University, 
Bunkyo-ku, Tokyo,
112--8610, Japan. 
\\
${}^\diamond$ &
Research Division, Mitsubishi Tanabe Pharma Corporation, 
Toda-shi, Saitama, 376--8515, Japan. 
\\
\end{tabular}
\end{center}

\begin{abstract}

\textbf{Background: }
Glomerulus detection is a key step in histopathological evaluation of microscopy images of kidneys. However, the task of automatic detection of glomeruli poses challenges due to the disparity in sizes and shapes of glomeruli in renal sections. Moreover, extensive variations of their intensities due to heterogeneity in immunohistochemistry staining are also encountered.\\  
\noindent Despite being widely recognized as a powerful descriptor for general object detection, the rectangular histogram of oriented gradients (Rectangular HOG) suffers from many false positives due to the aforementioned difficulties in the context of glomerulus detection. 

\textbf{Results: }
A new descriptor referred to as Segmental HOG is developed to perform a comprehensive detection of hundreds of glomeruli in images of whole kidney sections. The new descriptor possesses flexible blocks that can be adaptively fitted to input images to acquire robustness to deformations of glomeruli. Moreover, the novel segmentation technique employed herewith generates high quality segmentation outputs and the algorithm is assured to converge to an optimal solution. Consequently, experiments using real world image data reveal that Segmental HOG achieves significant improvements in detection performance compared to Rectangular HOG. 

\textbf{Conclusion: }
The proposed descriptor and method for glomeruli detection present promising results and is expected to be useful in pathological evaluation.

\end{abstract}

\section{Background}
%
Renal glomeruli provide a filtration barrier that retains higher molecular weight proteins in blood circulation. In various renal diseases, damage of the glomerular filtration barrier can be observed as protein leakage into urine, known as proteinuria. Therefore, the pathological changes in renal glomeruli of animal disease models can provide important information in screening compounds that target such diseases.

Our goal is to perform high-throughput detection of glomeruli in huge microscopy images of animal disease models, whose sizes run up to the order of $10^{8}$ pixels. While existing studies about automatic analysis of glomeruli in microscopy images of kidneys are present~\cite{ZhaHu08a,MaZhaHu09a}, target images in these works are from human biopsy samples with relatively small sizes, and are not suitable for our purpose.

Compared to general object detection tasks, there are two particular obstacles in the case of glomerulus detection. The first obstacle arises from the non-rigid sizes and shapes of the targets within the images. Indeed, the sizes of glomeruli are stable \textit{in vivo}, although they swell in unfavorable situations, e.g. hypertension~\cite{HugPueHoy13}
and diabetes~\cite{RasLauTho05}, to some degree. Also, the sizes of glomeruli in a whole-kidney-section image could vary depending on which part of a glomerulus the cross section passes through. The shapes of the glomeruli are almost spherical, making the boundaries circular. To obtain the boundaries, one might try to fit an ellipse to each glomerulus. However, this approach yields large estimation errors because each glomerulus is deformed to some extent.

The second difficulty arising in glomerulus detection task is the high variation of the intensities. In histological evaluation, immunohistochemistry is usually used to demonstrate the distribution and location of proteins in sections. In our target images, sections are immunostained for desmin, a known glomerular injury marker. As a result, some glomeruli are stained, and some are not. Since many glomeruli are partly stained, yielding heterogeneously stained glomeruli, detection is more complicated. Furthermore, the stained tissues in the kidneys are not only from glomeruli but also from other tissues such as blood vessels.

To check the existence of a glomerulus at each location in a whole-kidney-section image, the sliding window technique~\cite{MunGav06,MajBerMal08a,PapPog00a,VioJon04a} is employed. Using this procedure, a frame goes over the input image to check at every possible location whether the target object exists, then, a descriptor of the sub-image is extracted. 

Rectangular HOG (R-HOG)~\cite{Dalal05}, a widely used and recognized efficient descriptor for object detection in the field of computer vision, is a potentially suitable candidate descriptor for glomeruli. It has the capacity to capture the information of the magnitudes of the gradients in the image. Therefore, it is robust to the change in intensities caused by the heterogeneity of the stained levels. Glomeruli are known to be composed of tightly packed cells, resulting to high gradients in images. Thus, a natural approach would be to use the magnitudes of these gradients as features for the glomeruli. While we have also previously attempted to directly exploit this attribute, we have found the detection performance to be poor, resulting to many false positives and low recall. In addition to the magnitudes of gradients, their directions are also important to distinguish glomeruli from the other tissues. Using R-HOG descriptors obtained from both the magnitudes and the directions of the gradients, glomeruli detection performance results to recall values high enough to be useful for pathological evaluation. However, it appeared that R-HOG still suffers from a considerable amount of false positives~\cite{HirRel1x,KakOkaHir14a,KakOkaFuj15}.

The high number of false positives from the previous studies~\cite{HirRel1x,KakOkaHir14a,KakOkaFuj15} can be ascribed to the condition that the standard HOG such as the R-HOG has a rigid block division. Due to this rigidity, there are instances when a block is inside the glomerular area, and outside on another. Thus, extracted features from each block contain large variances, and robustness to the deformations of glomeruli is lost. Although there are several other known local descriptors such as scale-invariant feature transform (SIFT) image features~\cite{Lowe04a}, Haar-like features~\cite{VioJon04a}, and local binary patterns (LBP)~\cite{AhoHadPie06},  these do not possess a solution to be robust to the deformation of glomeruli for similar reasons.

In this study, we introduce a flexible block division to the HOG descriptor to improve the detection performance and reduce the number of false positives. A new feature, which we refer to as the Segmental HOG (S-HOG) descriptor, is proposed for glomerulus detection. The block division of S-HOG is based on the estimated boundary of the glomerulus that is obtained via a segmentation algorithm, which is also developed in this work. This renders the division of blocks to be more adaptable than the rigid block division of R-HOG, and allows feature vectors to clearly differentiate between the inside and the outside of the glomerulus. Moreover, since blocks are always within the glomerular area, gradient information in the same block between two glomeruli is expected to be more similar. Experiments conducted reveal that the number of false positives was halved, keeping almost all true positives when using S-HOG compared to the R-HOG.

\subsection*{Related Works}
Segmentation is an important step to extract the S-HOG descriptors. 
Recent works on segmentation of glomeruli has been sparse \cite{ZhaHu08a,MaZhaHu09a}. Nevertheless, there has been some research regarding
segmentation of specific organs in general biomedical images, including region growing~\cite{JiaHeFan13}, level set method~\cite{Cremers:2007}, and active contour 
model~\cite{KasWitTer88,WanCheBas09,Xie10a}. Majority of them are semi-automatic and require users\text{'} intervention, possess no guarantee of optimality~\cite{KasWitTer88,WanCheBas09,Xie10a}, and are highly dependent on the initial solution provided by users as input. On the other hand, the segmentation algorithm developed in this study is ensured theoretically to obtain the optimal solution, producing high quality segmentation. In addition to the above-mentioned, most recent attempts include using deep learning~\cite{Ciresan12a}. Deep learning typically requires great computational and time resources, whereas the proposed algorithm can work even on a standard personal computer or a laptop.

The algorithm developed by Kvarnstr\"{o}m et al.~\cite{Kvarnstrom:08} is relevant to the proposed segmentation technique. Their algorithm for cell contour recognition is based on a dynamic program, where  they first estimated the cell centers and constructed a ray from the center to each of $m$ directions, where $m=32$. Then they computed the boundary likeliness at $n$ points on each ray, where they set $n=30$. Their algorithm finds a smooth contour by taking a point on each ray to connect them. To get a closed contour, they presented two algorithms. The first algorithm poses $n$ sub-problems where in each sub-problem, the initial point and the endpoint are the same. We shall refer to this algorithm here as the \emph{exhaustive dynamic program} (EDP). Their second algorithm is a heuristic method that is faster than the first scheme, but possesses no guarantee for global optimality. In this study, we developed a new segmentation algorithm which we will refer to as \emph{divide \& conquer dynamic program} (DCDP). Compared to Kvarnstr\"{o}m et al's algorithms~\cite{Kvarnstrom:08}, the advantages of the DCDP algorithm are as follows:  
\begin{itemize}
\item DCDP is much faster than EDP, yet an exact optimal solution is always obtained. 
\item The boundary likeliness function is trained with a machine learning technique to precisely estimate boundaries of glomeruli.  
\end{itemize}


\section{Methods}

\begin{figure}[h]
\begin{center}
\begin{tabular}{p{0.22\textwidth}p{0.22\textwidth}}
\multicolumn{2}{c}{\includegraphics[width=0.45\textwidth]{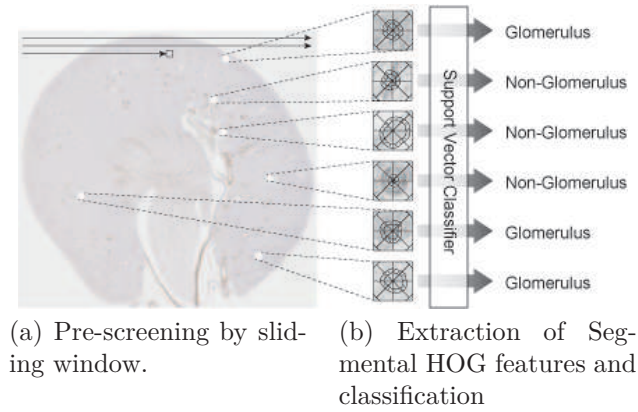}}
\\
(a) Pre-screening by sliding window. 
&
(b) Extraction of Segmental HOG features and classification
\\
\end{tabular}
\end{center}
\caption{\textbf{Flow of Our Method.}
In this study, a new descriptor, Segmental HOG, is developed
to detect glomeruli in huge microscopy images. SVM is
combined with Segmental HOG to classify candidates of
glomeruli that passed the pre-screening stage.
\label{fig:demo61-03-flow}
}
\end{figure}

In the proposed method, a new descriptor, S-HOG, is
introduced to detect glomeruli in kidney microscopic images.
Segmentation of glomeruli is needed to extract the S-HOG
descriptor. For fast exhaustive detection of glomeruli from
large microscopic images, pre-screening is performed with
R-HOG which does not require prior segmentation.
The proposed method consists of the three stages
(Figure~\ref{fig:demo61-03-flow}):
\begin{itemize}
\item pre-screening stage, 
\item segmentation stage, and
\item classification stage. 
\end{itemize}
In each stage, a support vector machine (SVM) \cite{SchSmo02,Shawe-Taylor04book} 
is used with a different type of HOG descriptor, resulting in three SVMs in total.
To obtain the S-HOG descriptor, we perform segmentation of glomeruli from the 
sub-images that passed the pre-screening (Figure~\ref{fig:demo61-03-flow}).  
Hereinafter, we present the details of each stage, and discuss how training datasets for each SVM are constructed and the materials used in the experiments at the end of this section.

\subsection*{Pre-Screening}\label{ss:pre-screening}
In the pre-screening stage, candidate glomeruli are detected from a kidney microscopy image using the sliding window technique. The window size is set to $200\times 200$ in our experiments. R-HOG features, which are 512-dimensional vectors based on our selected parameter values, are extracted and judged by SVM, and non-maximal suppression is then performed to obtain candidate glomeruli.

\subsection*{Segmentation}\label{ss:segmentation}
\begin{figure*}[h]
\begin{center}
\begin{tabular}{ll}
(a) & (b) 
\\
\includegraphics[width=0.4\textwidth]{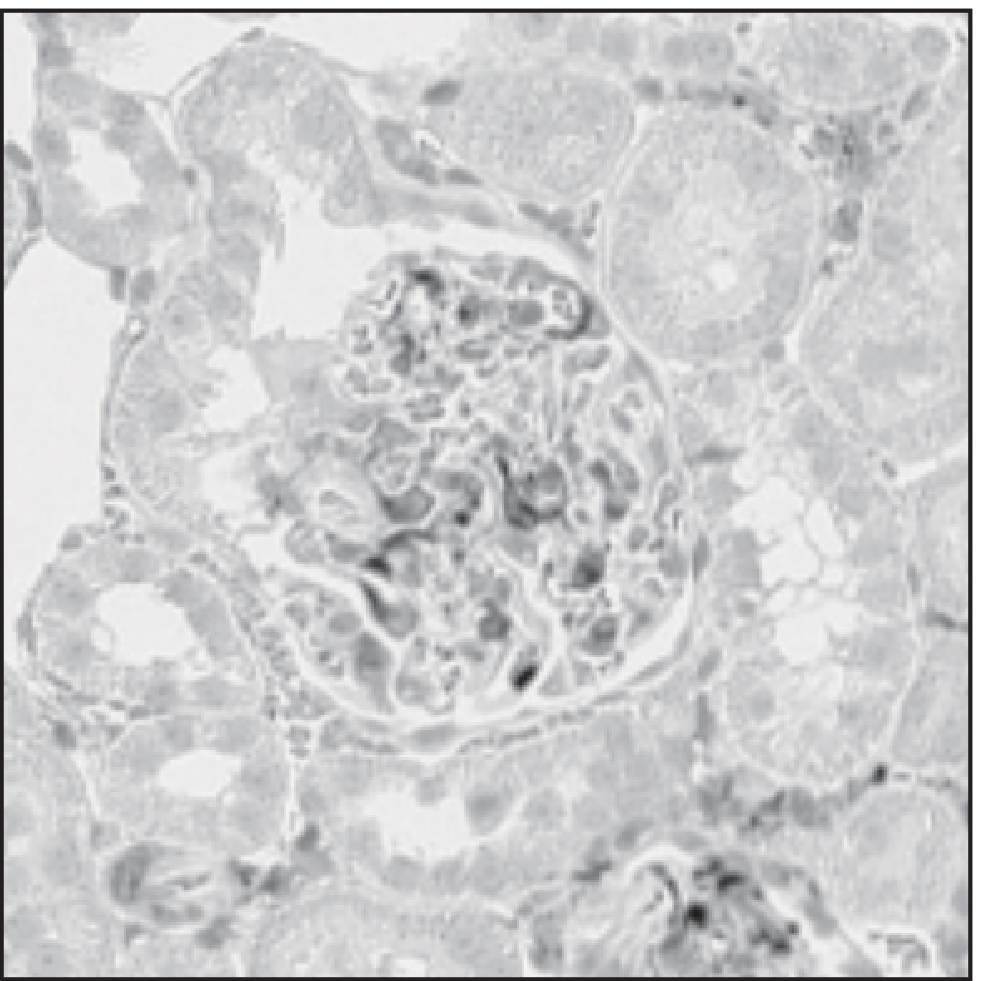}
&
\includegraphics[width=0.4\textwidth]{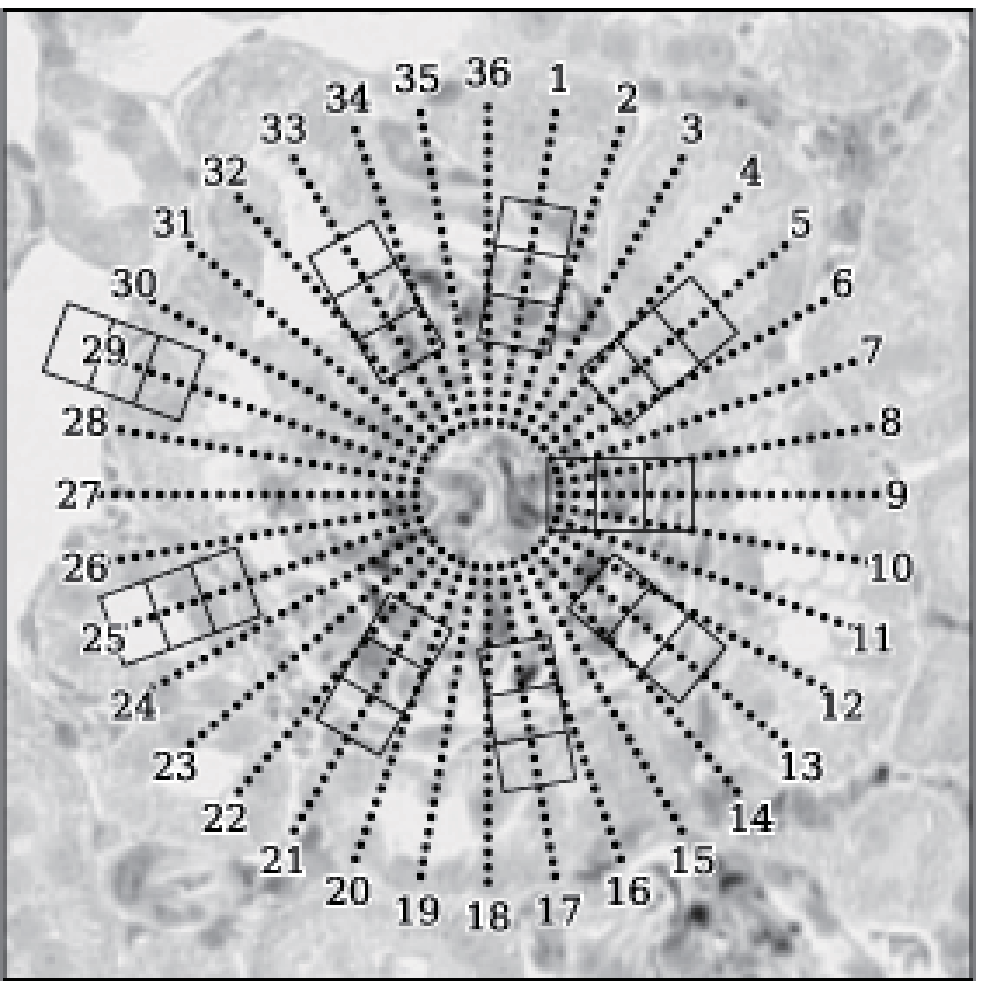}
\\
\end{tabular}
\end{center}
\caption{\textbf{Candidate Glomerulus and Line Segments.} 
S-HOG is based on the boundary of the objects of interest.
Suppose the boundary of a candidate glomerulus (Panel(a)) is
to be located. Boundary likeliness is computed at every
point on $m(=36)$ line segments placed uniformly in all $m$
directions (Panel(b)). The boundary likeliness is computed 
at $n$ points on each line segment. 
The $n$ locations are depicted with dots in Panel (b). 
\label{fig:demo61-03-inp}
}
\end{figure*}

Segmentation of glomeruli is performed on sub-images that passed the pre-screening. In the segmentation 
algorithm, the boundary of a glomerulus is represented by an
$m$-sided polygon whose $m$ vertices are restricted to lie
on $m$ line segments, respectively. The $m$ line segments
are placed uniformly, as outlined by the dotted lines 
in Figure~\ref{fig:demo61-03-inp}b 
for the case where $m=36$. To
determine the location of the vertex on each line segment,
the sliding window technique is employed again.
\footnote{Sliding windows are used in both pre-screening and 
segmentation.} 
The window sweeps
through the line segment and computes the \emph{boundary
likeliness} at $n$ locations on the line segments. 
In Figure~\ref{fig:demo61-03-inp}b,  the boundary likeliness $L_{i}$ $(i=1,\ldots,m)$ is computed at every dotted location.  How $L_{i}$ is computed is discussed at the end of this {\sffamily Segmentation} subsection.
We set the length of the line segment to $63$ pixels, where the endpoint closest to the center of the image is $17$ pixels away from the center. 
The length between adjacent dots along a line segment is equal to $3$ pixels, resulting to $n=22$ dots on each line segment. 
In total, the boundary likeliness is computed at $mn(=36\times 22=792)$ locations. 
To
determine the vertices of the $m$-sided polygon, one might
consider na\"{i}vely locating the points that achieve the highest
boundary likeliness on each line segment. However, this approach often yields an extremely zigzag boundary. 

To obtain a smoother boundary, we impose a constraint that
suppresses distant adjacent vertices. We then establish the
following maximization problem: 
\begin{equation}\label{eq:origprob-glomdp}
\begin{split}
\max\quad&
\sum_{i=1}^{m}L_{i}(p_{i}),\qquad
\text{wrt }\quad
p_{1},\dots,p_{m}\in\{1,\dots,n\},
\\
\text{subj to }\quad&
|p_{1}-p_{2}|\le\varsigma, \dots,
|p_{m-1}-p_{m}|\le\varsigma, \,
|p_{m}-p_{1}|\le\varsigma, 
\end{split}
\end{equation}
where
$L_{i}:\{1,\dots,n\}\to\bR$ denotes the boundary likeliness function
obtained by the sliding window technique, 
and $p_{i}\in\{1,\dots,n\}$
$(i=1,\dots,m)$ is a location on the $i$-th line
segment, where the line segment is discretized into $n$
points numbered with a natural number. 
For instance, when $p_{i}=1$, 
the $i$-th vertex is at the endpoint of the $i$-th line 
segment closest to the center, and 
the vertex can move from this endpoint to
the other endpoint with increasing value of $p_{i}$. 
$L_{i}(p)$ is the boundary likeliness at $p$-th location 
in the $i$-th line segment. 
The location $p_{i}$ on the $i$-th line segment is 
more likely to be the
boundary with larger $L_{i}(p_{i})$ value. In our experiments, we set $\varsigma=1$. 
We shall denote the objective function by $J(\vp)$, i.e., $J(\vp):=\sum_{i=1}^{m}L_{i}(p_{i})$. 


Kvarnstr\"{o}m et al. tackled a similar optimization problem, and proposed an algorithm which we called EDP. In this study, a faster algorithm named DCDP is developed. The two algorithms, EDP and DCDP, are detailed as follows.

\paragraph*{Exhaustive Dynamic Program (EDP).} 
Due to the last constraint $|p_{m}-p_{1}|\le\varsigma$, 
the standard dynamic program cannot directly solve 
optimization problem \eqref{eq:origprob-glomdp}. 
To make the problem tractable, EDP divides
problem \eqref{eq:origprob-glomdp} into $n$ sub-problems,
where the value of $p_{m}$ is fixed to one of
$\{1,\dots,n\}$ in each sub-problem. Once all
the $n$ sub-problems are solved, the optimal
solution of the original problem~\eqref{eq:origprob-glomdp} is 
obtained by taking the
solution that yields the largest objective value
among $n$ sub-problems. 
Each sub-problem
can be solved by a dynamic program that takes 
$O(nm\varsigma)$ computation. 
This algorithm computes an optimal 
solution of the $k$-th sub-problem, which is equivalent to 
the original problem~\eqref{eq:origprob-glomdp}
to which the constraint~$p_{m}=k$ is added. 
Hence, the maximization problem \eqref{eq:origprob-glomdp}
can be solved with $O(n^{2}m\varsigma)$ computation. 

The new algorithm DCDP also takes $O(n^{2}m\varsigma)$ computational time in worst case, although the new algorithm solves the same problem much faster than EDP, as presented in the 
{\sffamily Results} section.

\paragraph*{Divide \& Conquer Dynamic Program (DCDP).} 
Observe that problem \eqref{eq:origprob-glomdp} can be solved in $O(nm\varsigma)$ computational time 
by a dynamic program if one of the constraints $|p_{m}-p_{1}|\le\varsigma$ 
is disregarded. 
The idea to devise the new algorithm is based on the
following fact: Suppose $\vp_{0}^{\star}$ is an optimal solution that maximizes $J(\cdot)$ without the constraint $|p_{m}-p_{1}|\le\varsigma$. Then if $\vp_{0}^{\star}$ is a feasible solution for the original problem~\eqref{eq:origprob-glomdp}, it is also an optimal solution.

To express this idea mathematically, 
let us define
\begin{align*}
\cS(\cI) :=&
\big\{ \vp = \left(p_{1},\dots,p_{m}\right) \in\bN_{n}^{m}\,|
\\
&\,\,p_{m}\in\cI, \,
|p_{1}-p_{2}|\le\varsigma, \dots, 
|p_{m-1}-p_{m}|\le\varsigma, 
|p_{m}-p_{1}|\le\varsigma 
\big\}. 
\end{align*}
for $\cI\subseteq\bN_{n}$, 
where $\bN_{n} := \left\{1,\ldots,n\right\}$. Note that $\cS(\bN_{n})$ is the feasible region of 
the original problem \eqref{eq:origprob-glomdp}.
The goal of DCDP is to find an optimal solution
\begin{align*}
\vp^{\star} \in \argmax_{\vp\in\cS(\bN_{n})} J(\vp). 
\end{align*}
Dynamic program (DP) cannot solve this problem directly
due to the existence of the constraint $|p_{m}-p_{1}|\le\varsigma$. 
To use DP, we consider finding the maximizer of $J(\vp)$ from a
relaxed region~$\cS_{\text{L}}(\bN_{n})$ where $\cS_{\text{L}}(\cI)$ 
is defined as
\begin{align*}
\cS_{\text{L}}(\cI)
:=&
\big\{ \vp = \left(p_{1},\dots,p_{m}\right) \in\bN_{n}^{m}\,|
\\
&
\,\,
p_{m}\in\cI, \,
p_{1}\in\cI+\{-\varsigma,\dots,+\varsigma\}, \,
|p_{1}-p_{2}|\le\varsigma, \dots, 
|p_{m-1}-p_{m}|\le\varsigma \big\}, 
\end{align*}
where the operator $+$ denotes that for any two sets $\cI$ and $\cJ$, $\cI+\cJ:=\{i+j\,|\,i\in\cI,\, j\in\cJ\}$. Note that $\cS(\cI)\subseteq\cS_{\text{L}}(\cI)$. 
The strategy of DCDP is 
to first find the solution of the relaxed problem, 
\begin{align*}
\vp_{0}^{\star} \in \argmax_{\vp\in\cS_{\text{L}}(\bN_{n})}J(\vp)
\end{align*}
and then check the feasibility: if 
$\vp_{0}^{\star}\in\cS(\bN_{n})$, then 
$\vp_{0}^{\star}$ is the optimal solution of 
the original problem~\eqref{eq:origprob-glomdp}. If 
$\vp_{0}^{\star}\not\in\cS(\bN_{n})$, then 
the set $\bN_{n}$ is divided into $\cI_{1}$ and $\cI_{2}$ 
(i.e. $\cI_{1}\cup\cI_{2} = \bN_{n}$), and 
the following two sub-problems are solved: 
\begin{align*}
\vp_{1}^{\star} \in\argmax_{\vp\in\cS(\cI_{1})}J(\vp)
\qquad\text{and}\qquad
\vp_{2}^{\star} \in\argmax_{\vp\in\cS(\cI_{2})}J(\vp). 
\end{align*}
Notice that the original feasible region $\cS(\bN_{n})$ 
is the sum of the two regions, $\cS(\cI_{1})$ and $\cS(\cI_{2})$. 
Therefore, we can take either of the two solutions, 
$\vp_{1}^{\star}$ and $\vp_{2}^{\star}$, which has the 
larger objective value.  DCDP employs
a divide and conquer approach that repeatedly applies 
the above strategy to sub-problems.  
The basic approach of DCDP is summarized in Algorithm~\ref{algo:circopt1}. 
Invoking the function $\textsc{DCDP\_Basic}(\bN_{n})$ 
yields the optimal solution of the original problem. 
Here, the function $(\cI_{1},\cI_{2}) := \textsc{Split}(\cI_{0})$ 
divides the set $\cI_{0}$ into two exclusive non-empty subsets, 
$\cI_{1}$ and $\cI_{2}$. 

\begin{algorithm}[th!]
\caption{$\vp_{0}=\textsc{DCDP\_Basic}(\cI_{0})$. 
\label{algo:circopt1}}
\begin{algorithmic}[1]
{\normalsize
\REQUIRE $\cI_{0}\subseteq\bN_{n}$. 
\ENSURE $\vp_{0} \in\argmax_{\vp\in\cS(\cI_{0})}J(\vp)$. 
\STATE $\vp_{0}\in\argmax_{\vp\in\cS_{\text{L}}(\cI_{0})}J(\vp)$; 
\STATE \textbf{if }$\vp_{0}\in\cS(\cI_{0})$, \textbf{then return}; 
\STATE $(\cI_{1},\cI_{2}) := \textsc{Split}(\cI_{0})$; 
\STATE $\vp_{1} := \textsc{DCDP\_Basic}(\cI_{1})$;  
\STATE $\vp_{2} := \textsc{DCDP\_Basic}(\cI_{2})$;  
\STATE $i^{\star} \in \argmax_{i\in\{1,2\}}J(\vp_{i})$; 
\STATE $\vp_{0} := \vp_{i^{\star}}$;
} 
\end{algorithmic}
\end{algorithm}
The first step $\vp_{0}\in\argmax_{\vp\in\cS_{\text{L}}(\cI_{0})}J(\vp)$ can be performed in $O(nm\varsigma)$ computational time. An instance of the dynamic program is given in Algorithm~\ref{algo:subprob-glomdp}. Note that $\vp_{0}\in\cS(\cI_{0})$ is always ensured if the cardinality of $\cI_{0}$ is one since the relaxed region is reduced to the unrelaxed region (i.e. $\cS(\{h\}) = \cS_{\text{L}}(\{h\})$). The function \textsc{DCDP\_Basic} is invoked, at most, $(2n-1)$ times.  This implies that the computational time in worst case is $O(n^{2}m\varsigma)$. As will be shown in the {\sffamily Results} section, we empirically found that the number of invoking the function recursively is much smaller than $(2n-1)$.    

To accelerate the DCDP algorithm, pruning steps are added. For the implementation of $\textsc{Split}(\cI_{0})$, we considered three schemes: 
\emph{Half Split}, 
\emph{Max Split}, and \emph{Adap Split}. The pruning steps and the three schemes of $\textsc{Split}(\cI_{0})$ are detailed in Sections~\ref{ss:pruning} and \ref{ss:splitting-schemes}, as well as the resulting accelerated DCDP algorithm.  
A mathematical proof using case analysis showing that 
our algorithm always obtains an optimal solution is also given 
in Section~\ref{ss:proof-of-lem-opt-dcdp}. 

\begin{algorithm}[tbh]
\caption{$O(nm\varsigma)$ Dynamic Program for 
$\max_{\vp^{\star}\in\cS_{\text{L}}(\cI)}J(\vp)$
\label{algo:subprob-glomdp}}
\begin{algorithmic}[1]
{\normalsize
\REQUIRE $\cI \subseteq\bN_{n}$. 
\ENSURE $\vp\in\argmax_{\vp\in\cS_{\text{L}}(\cI)}J(\vp)$
\STATE
Initialize all entries in the $n\times m$ matrix~$Q$ with $-\infty$. 
\FOR{$j\in \cI + [-\varsigma,+\varsigma]\cap\bN_{n}$}
\STATE
$Q(j,1) := L_{1}(j);$ 
\ENDFOR
\FOR{$t=2,\dots,m$}
\STATE $\text{if }t < m, \textbf{then } \cI_{t}:=\bN_{n} \text{else }\cI_{t}:=\cI$; 
\FOR{$i\in\cI_{t}$}
\STATE
$j_{\star} := \argmax
\{ Q(j,t-1) \,\,|\,\,
{j\in[i-\varsigma,i+\varsigma]\cap\bN_{n}}\};$
\STATE
$Q(i,t):=L_{t}(i)+Q(j_{\star},t-1);$ 
$\,\,P(i,t):=j_{\star};$
\ENDFOR
\ENDFOR
\STATE
$p^{\star}_{m}\in\argmax_{i\in\cI} Q(i,m)$; 
\STATE
$t:=m-1$; 
\WHILE{$t\ge 1$}
\STATE 
$p^{\star}_{t}:=P(p^{\star}_{t+1},t+1)$; $\quad t:=t-1$; 
\ENDWHILE
}
\end{algorithmic}
\end{algorithm}

\paragraph*{Computation of Boundary Likeliness~$L_{i}(\cdot)$.}
The sliding window technique is employed in order to determine the vertices of the $m$-sided polygon described above. The window size in this stage is set to $30\times 15$ pixels, and the windows sweep through the line segments (Figure~\ref{fig:demo61-03-inp}b). Each time the sliding window moves, a feature descriptor is computed from the window and is applied with Linear SVM to compute the SVM score, which is what we refer to as the boundary likeliness $L_{i}(\cdot)$.
The SVM scores of $m$ vertices, $L_{i}(1),\dots,L_{i}(n)$, are then obtained for $i=1,\dots,m$, and integrated in the maximization problem~\eqref{eq:origprob-glomdp}.

The HOG feature is adopted as the descriptor to compute the boundary likeliness. Each window is divided into three blocks as shown in Figure~\ref{fig:demo61-03-inp}b. This division design is from an observation that some glomeruli are surrounded with a thick Bowman\text{'}s capsule, and the middle block is expected to capture this glomerular capsule. The statistics of nine discretely oriented gradients are computed in each block, producing a 27-dimensional feature vector.

\subsection*{Classification with S-HOG descriptor}\label{ss:classify}

\begin{figure}[h]
\begin{center}
\begin{tabular}{cc}
\includegraphics[width=0.15\textwidth]{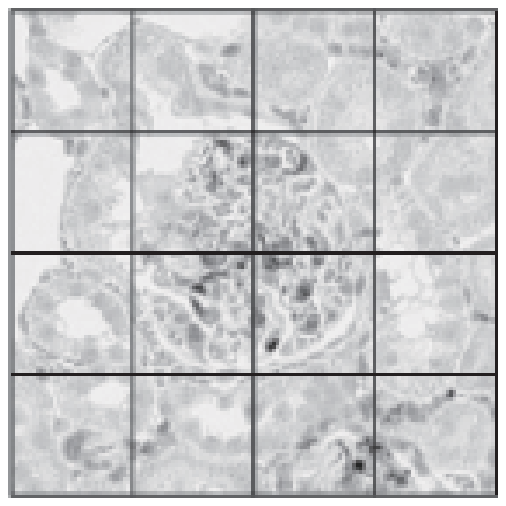}
&
\includegraphics[width=0.3\textwidth]{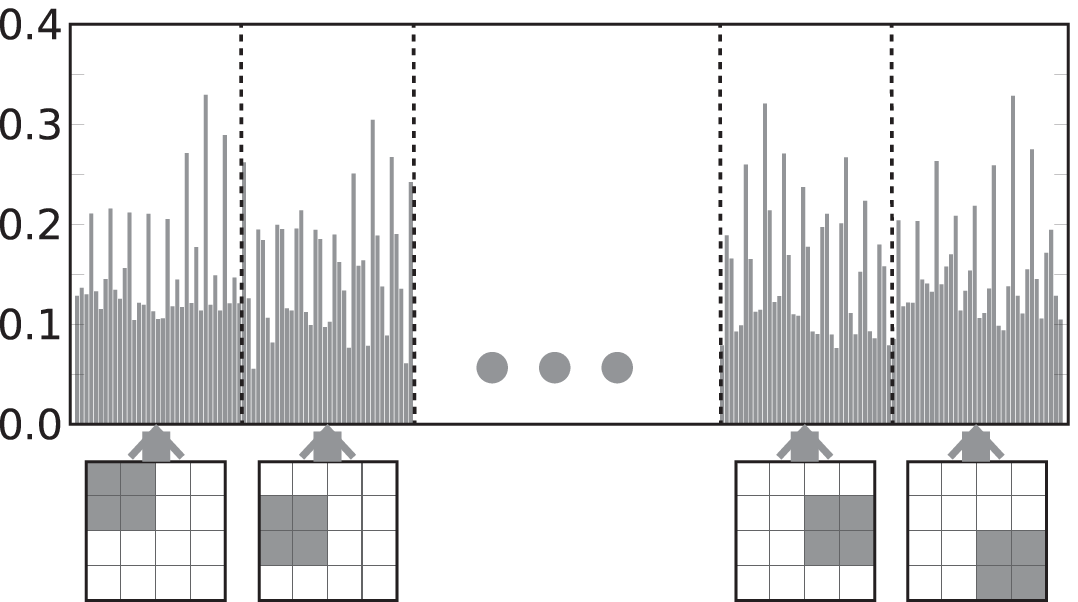}
\\
(a) Blocks of R-HOG & (b) Feature Vector of R-HOG
\\
\\
\includegraphics[width=0.15\textwidth]{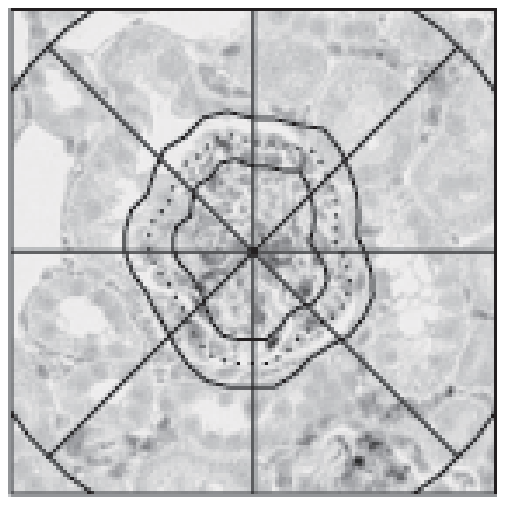}
&
\includegraphics[width=0.3\textwidth]{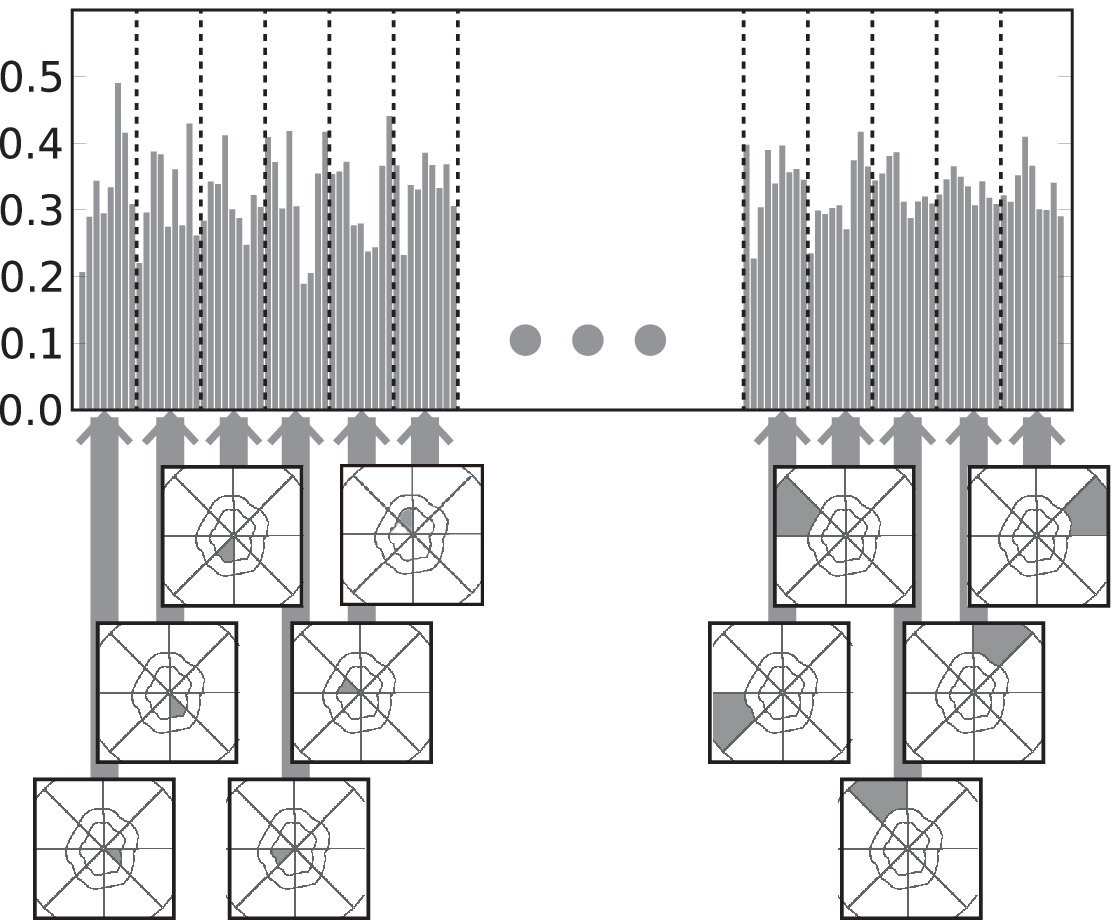}
\\
(c) Blocks of S-HOG & (d) Feature Vector of S-HOG
\\
\end{tabular}
\end{center}
\caption{\textbf{Rectangular HOG and Segmental HOG.}
R-HOG has been used for object detection in a wide variety
of applications. R-HOG is the concatenation of statistics in
each block in a grid dividing a rectangular region. On the other hand, the
blocks of the proposed descriptor, S-HOG, are based on the
segmentation of the object of interest.
\label{fig:demo61-05-twohogs}
}
\end{figure}

Candidate glomeruli obtained via pre-screening are
classified using the proposed S-HOG descriptor. S-HOG
exploits the glomerulus boundary located in the segmentation
stage to generate 24 non-overlapping blocks, as shown in 
Figure~\ref{fig:demo61-05-twohogs}c. 

Various types of glomeruli are contained in kidney microscopy images, some of them surrounded with a thick Bowman\text{'}s capsule. To effectively exploit this characteristic, the circle containing a candidate glomerulus is divided into three zones: Inner zone, middle zone, and outer zone. We divide the circle into eight disjoint sectors, and take the intersection of each zone and each sector to get the 24 non-overlapping blocks (Figure~\ref{fig:demo61-05-twohogs}c), and gradients are then histogrammed for each block (Figure~\ref{fig:demo61-05-twohogs}d). In our experiments, we employed nine discretized oriented gradients, and SVM is applied to S-HOG feature vectors to discriminate between glomeruli and non-glomeruli.

\subsection*{Construction of Training Data}\label{ss:learning}
A total of three linear SVMs are used, one for the pre-screening, segmentation, and classification stage, respectively. A training dataset is required for each of the three SVMs. Details on the construction of each training data set are given subsequently.

\paragraph*{Training Data for Pre-Screening Stage.}
Each example in the training data for pre-screening is a
$200\times 200$ sub-image. A positive example contains a
glomerulus in the center of the sub-image, while a negative example
does not. To gather these samples, the locations
of glomeruli within the whole-kidney-section images used for training are first annotated manually. 
Small glomeruli whose diameters are under
50 pixels were ignored. Positive examples are the sub-images
from $200\times 200$ bounding boxes containing an annotated
glomerulus in the center. Negative examples are $200\times
200$ sub-images picked from random locations in the kidney
microscopy images. From each sample, a 512-dimensional
R-HOG descriptor is extracted.

\paragraph*{Training Data for Segmentation Stage.}
As described in the {\sffamily Segmentation} subsection, the boundary likeliness is
computed in every position on the $m$ line segments. This
boundary likeliness is the SVM score. The position lying on
the true boundary of a glomerulus is considered as a positive example for
the SVM, and the other positions are negative examples. To
construct the training data for segmentation, the
positive sub-images in the training dataset for
pre-screening are reused.

\paragraph*{Training Data for Classification Stage.}
Examples in the training data for pre-screening are used
again for training in the classification stage,
but with a different set of features extracted via S-HOG. For each training data sample, 
the previously described segmentation algorithm estimates the boundary of the glomerulus. 
Based on the estimated boundary, the statistics of oriented gradients are
computed to obtain S-HOG feature vectors. This procedure is
done for both positive and negative examples,
even though negative examples do not contain a glomerulus.

\subsection*{Materials}\label{s:mate}
The images used in the present study had been generated in
a previous study~\cite{HirRel1x}, and only
an overview is given in this subsection.

Male 6-week-old SD and SDT rats were purchased from CLEA,
Inc. (Tokyo, Japan) and were housed with a 12-h light-dark
cycle and free access to water and chow.

Five SD and SDT Rats at 16 and 24 weeks of age were
euthanized under ether anesthesia. Their kidneys were
removed and immediately fixed in 10\% neutralized buffered
formalin. The formalin-fixed kidneys were embedded in
paraffin. For immunohistochemistry, kidney paraffin sections
were deparaffinized and incubated overnight at 4$^{\circ}$C with
anti-desmin mouse monoclonal antibody (Dako, Glostrup,
Denmark) followed by horseradish peroxidase-conjugated
secondary antibody (anti-mouse immunoglobulin goat
polyclonal antibody; Nichirei, Tokyo, Japan). The sections
were colorized brown with 3,3'-diaminobenzidine. Whole slide
images of the sections were obtained with Aperio Scan Scope
XT (Leica Microsystems, Wetzlar, Germany). All animal
experiments were performed in accordance with the Act on
Welfare and Management of Animals and the institutional
guidelines, and approved by the institutional Committee of
Animal Experiments of New Drug Development Research Center
Inc. (Hokkaido, Japan).

\section{Results and Discussion}\label{s:exp}
%
%
In this section, the detection performance is demonstrated
by showing the experimental comparisons between S-HOG and 
R-HOG~\cite{HirRel1x,KakOkaHir14a}. 

A set of 20 whole-kidney-section images is 
used in the experiments. 
The dataset is the same as the one used by \cite{HirRel1x}. 
The image sizes are $9,849\times 10,944$ pixels in average. Each
image is from one of four groups: 16-week-age SD rat,
16-week-age SDT rat, 24-week-age SD rat, and 24-week-age SDT
rat. Henceforth, for simplicity, we will refer to them as 16SD, 16SDT, 24SD, 24SDT,
respectively, each group containing five images. 
For performance evaluation, we annotated every glomerulus in the images manually. We divided the image set into five subsets: Set A, Set B, Set C, Set D, and Set E. Each subset consists of a 16SD image, a 16SDT image, a 24SD image, and a 24SDT image. For assessment of detection performance, the position of every glomerulus in the images is annotated and, for evaluation of segmentation performance, the areas of glomeruli in Set A and Set B are located manually using a graphics software.

As described in the previous section, our method has three stages: pre-screening, segmentation, and classification. 
Each stage uses its own
SVM trained with a hyper-parameter $C$. In the
classification stage, a threshold $\theta$ is used to classify an
example; if the SVM score is over the threshold $\theta$,
the example is predicted as positive, otherwise, negative.
For the pre-screening and classification stages, Set A was used for training SVM, and Set B was exploited for determining the optimal combination of $(C,\theta)$. Sets C, D, and E were for performance evaluation. SVM for the segmentation stage provides us with the boundary likelihood function. The regularization parameter $C$ for the SVM is determined via the holdout method within Set A. Seventy percent of the glomeruli in Set A are randomly selected for training, and the rest is used for validation. The resulting parameter values were $(C,\theta)=(10,2)$ for pre-screening, $C=10$ for segmentation, and $(C,\theta)=(10,-1.5)$ for classification.

\subsection*{Detection Performance}\label{ss:detect-perf}
\begin{figure*}[h]
\begin{center}
\begin{tabular}{cc}
\includegraphics[width=0.4\textwidth]{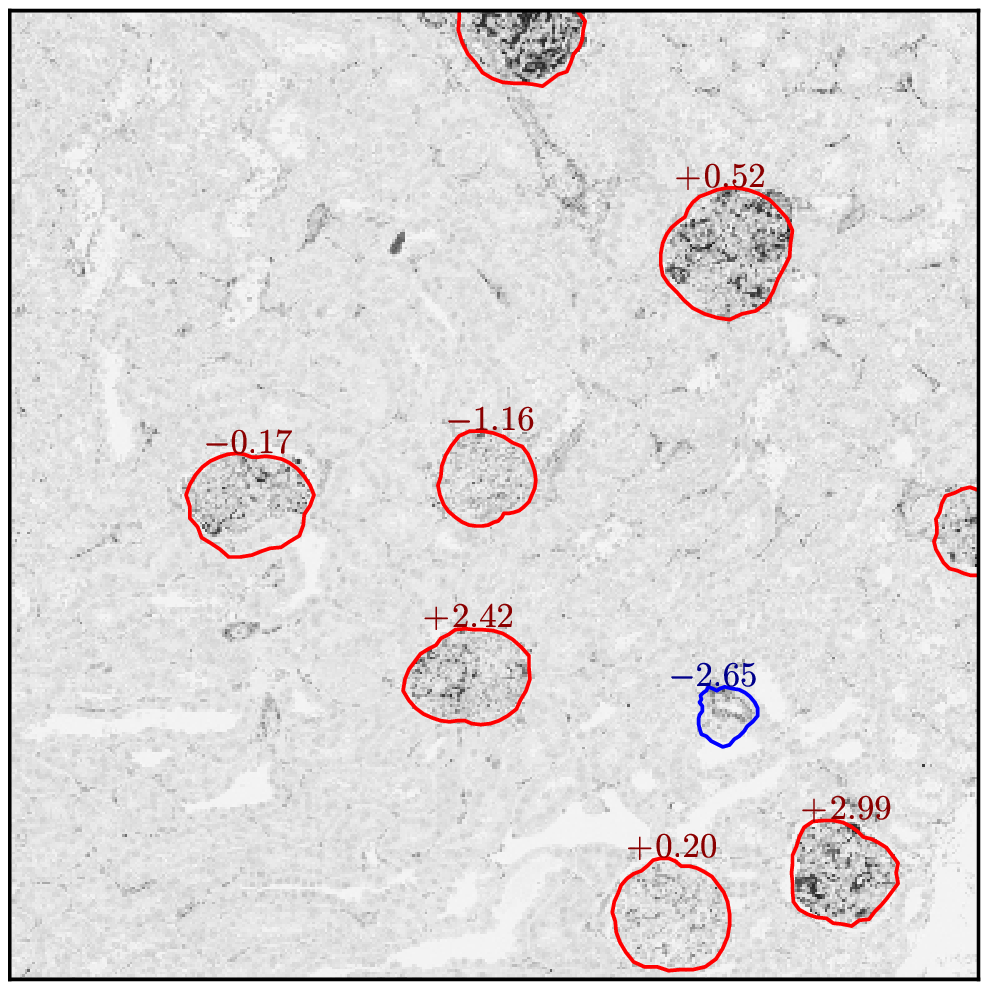} &
\includegraphics[width=0.4\textwidth]{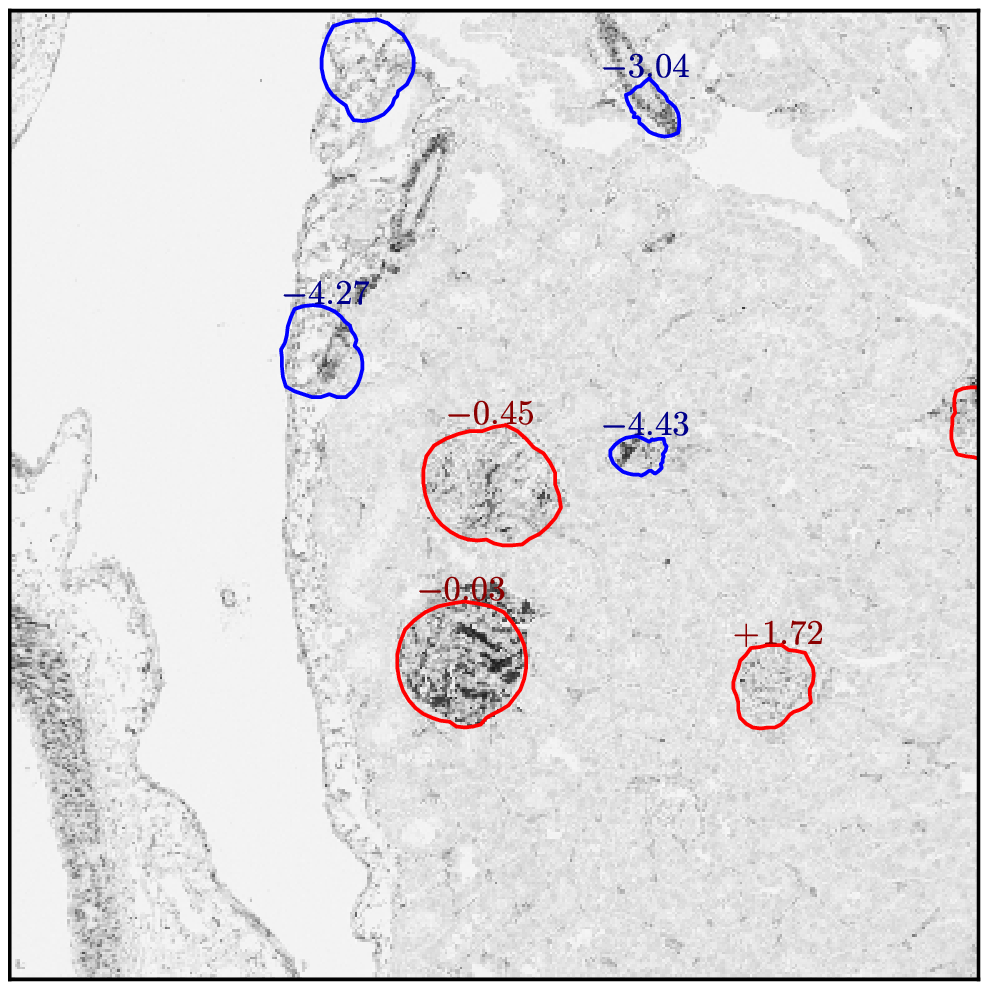}
\end{tabular}
\end{center}
\caption{\textbf{Examples of Detected Glomeruli.}
The estimated boundaries of the glomeruli are depicted with
red rings. The areas surrounded with blue rings are passed
through the pre-screening stage, but removed in the
classification stage. The numbers are the SVM scores resulting
from the classification stage. The areas with SVM scores over
$\theta=-1.5$ are classified as a glomerulus. 
It can be observed that false positives such as vessels detected 
in the pre-screening stage were successfully removed in the 
classification stage. 
\label{fig:demo81-06-demo}
}
\end{figure*}

%
%
Figure~\ref{fig:demo81-06-demo} 
illustrates examples of detected glomeruli. In the
two images, the candidate glomeruli passed through
pre-screening are depicted with rings that represent the
boundaries estimated in the segmentation stage. The numbers
printed above the rings are the scores produced by SVM in the
classification stage. Candidate glomeruli with SVM scores
below $\theta=-1.5$ are excluded from the final detection results.
The excluded candidates are depicted with blue rings, and the
remaining glomeruli with red rings. It can be observed
that non-glomerulus areas are excluded effectively,
whereas true glomeruli are estimated correctly.

For quantitative assessment of detection performance, 
true positives, false positives, and false negatives have to
be defined. True positive glomeruli (TPG) are identified as
correctly detected glomeruli, false positive glomeruli (FPG)
are wrongly detected glomeruli, and false negative glomeruli
(FNG) are the ones that could not be detected. From the
definitions of TPG, FPG, and FNG, we can compute for the
three widely used performance measures: F-measure, Precision, and Recall.
Precision is the ratio of TPG to detected glomeruli (i.e.
$\text{TPG}/(\text{TPG}+\text{FPG})$), 
Recall is the ratio of TPG to true
glomeruli (i.e. $\text{TPG}/(\text{TPG}+\text{FNG})$), and 
F-measure is the harmonic
mean of the Precision and the Recall.

\begin{figure}[h]
\begin{center}
\begin{tabular}{c}
\includegraphics[width=0.45\textwidth]{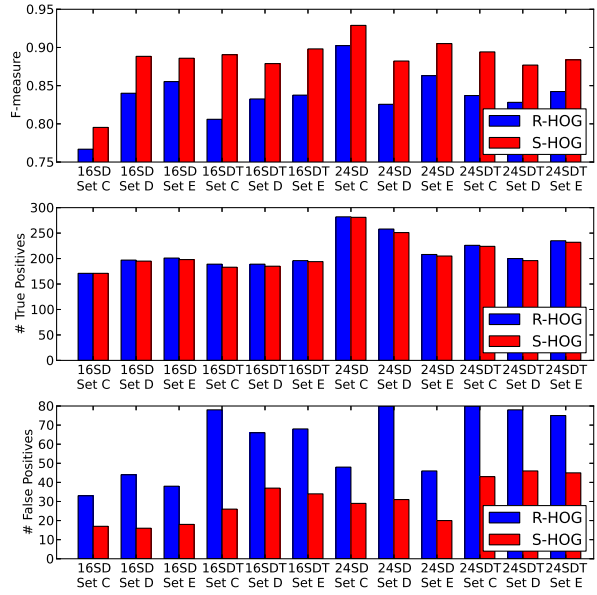}
\end{tabular}
\end{center}
\caption{\textbf{Detection Performances.}
The proposed descriptor, S-HOG, achieves evident 
improvement in F-measure compared to the existing descriptor, 
R-HOG. With small loss of true positives, S-HOG halves false positives
of R-HOG. (See subsection on {\sffamily Detection Performance} for details.) 
\label{fig:demo71-01-perf-twohogs}
}
\end{figure}

Figure~\ref{fig:demo71-01-perf-twohogs}
shows the plots of the F-measure, Precision, and Recall for each
testing image. S-HOG achieves an average of 0.866, 0.874, and 0.897
for F-measure, Precision, and Recall, respectively,
whereas R-HOG obtained 
0.838, 0.777, and 0.911, respectively. 
In applying detection methods to pathological evaluation, 
Precision is more important than Recall~\cite{KakOkaHir14a}, 
and in this study, S-HOG achieved considerably higher Precision at a small sacrifice of 
Recall. Two-sample t-test is
performed to assess the differences statistically. 
While no statistical difference of Recall can 
be detected (P-value is $3.47\cdot 10^{-1}$),
the differences among F-measure and Precision 
are significant (P-values is $1.34\cdot
10^{-3}$ and $3.75\cdot 10^{-5}$, respectively), 
%


\subsection*{Segmentation Performance}\label{ss:seg-perf}
\begin{figure*}[h]
\begin{center}
\begin{tabular}{ccc}
\includegraphics[width=0.3\textwidth]{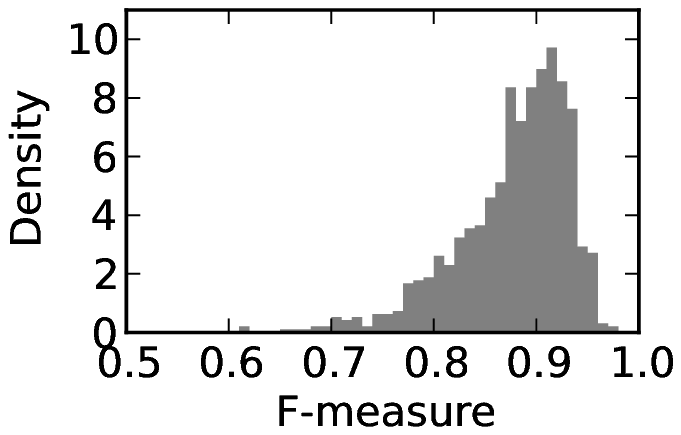}
&
\includegraphics[width=0.3\textwidth]{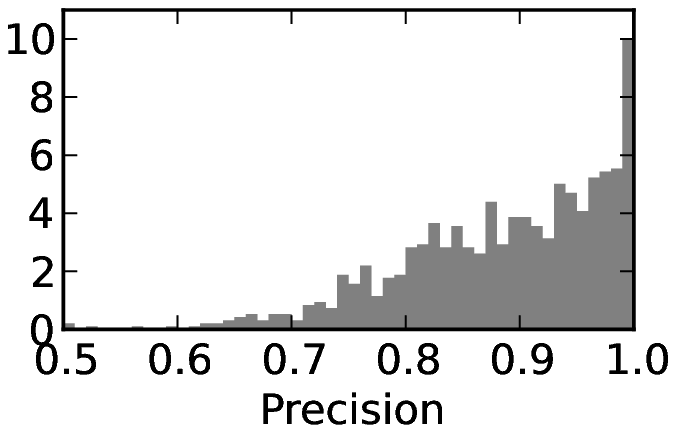}
&
\includegraphics[width=0.3\textwidth]{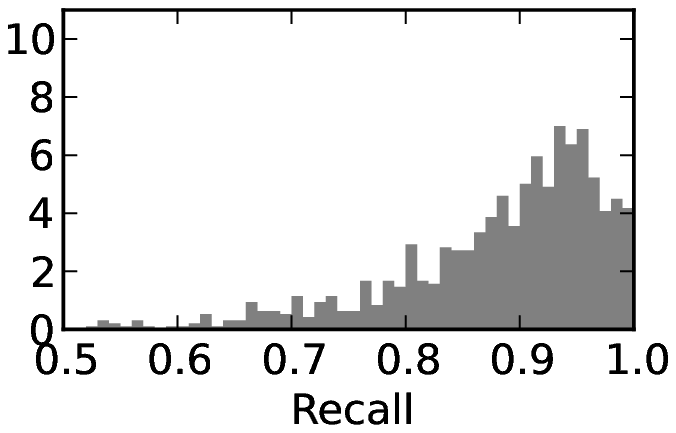}
\end{tabular}
\end{center}
\caption{\textbf{Segmentation Performances.}
The number of glomeruli are tallied to make a histogram with the 
F-measure, Precision, and Recall of the pixels based on 
comparison of true segmentation with estimated segmentation on the $x$-axes. 
(See {\sffamily Segmentation Performance} subsection for details.) 
\label{fig:demo71-02-seghist}
}
\end{figure*}

One advantage of the proposed method is that the boundaries
of the detected glomeruli can be obtained. These boundaries 
provide useful information for pathological evaluation~\cite{KakOkaHir14a}. 
Here we give a discussion of the performance of the segmentation algorithm. 
To quantify the accuracy of the estimated areas within the predicted boundaries, 993 annotated glomeruli in Set B were used. True positive area
(TPA), false positive area (FPA), and false negative area
(FNA) are defined as follows: TPA is the intersection of the
true area and estimated area; FPA is the relative complement
of the true area in the estimated area; FNA is the relative
complement of the estimated area in the true area. For each
glomerulus and its estimated area, F-measure, Precision, and
Recall can be obtained by counting the pixels in the TPA, FPA,
and FNA.
The histograms of the F-measure, Precision, and Recall are
plotted in Figure~\ref{fig:demo71-02-seghist}, 
where the frequency is normalized so
that the integral is one. Among the glomeruli, $90.1\%$ are
estimated to have F-measure over $0.8$, ensuring reliable assessment 
of medicinal effect for pharmaceutical development. 

\begin{figure}[ht]
\begin{center}
\begin{tabular}{ll}
(a) & (b)
\\
\includegraphics[width=0.45\textwidth]{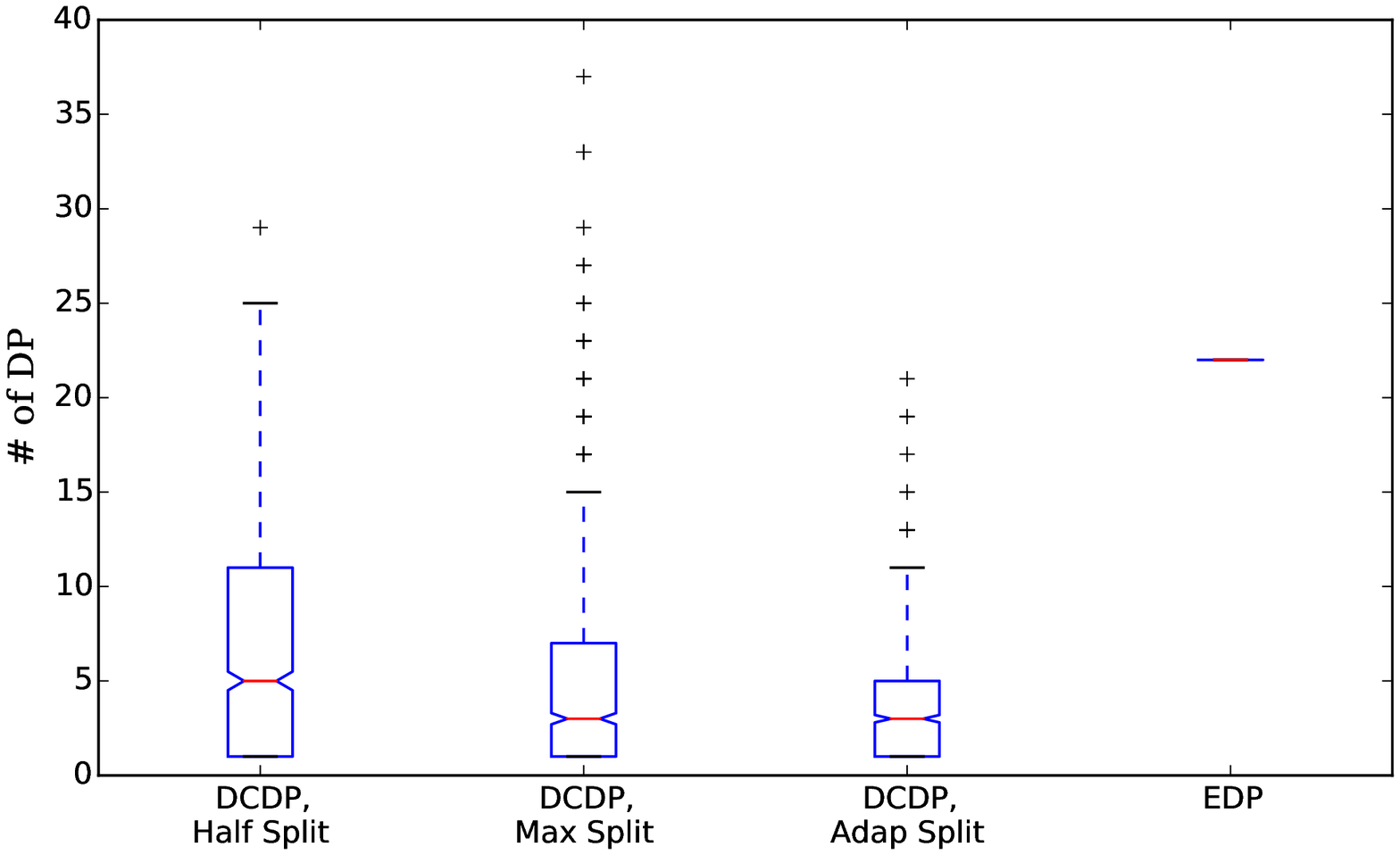}
&
\includegraphics[width=0.45\textwidth]{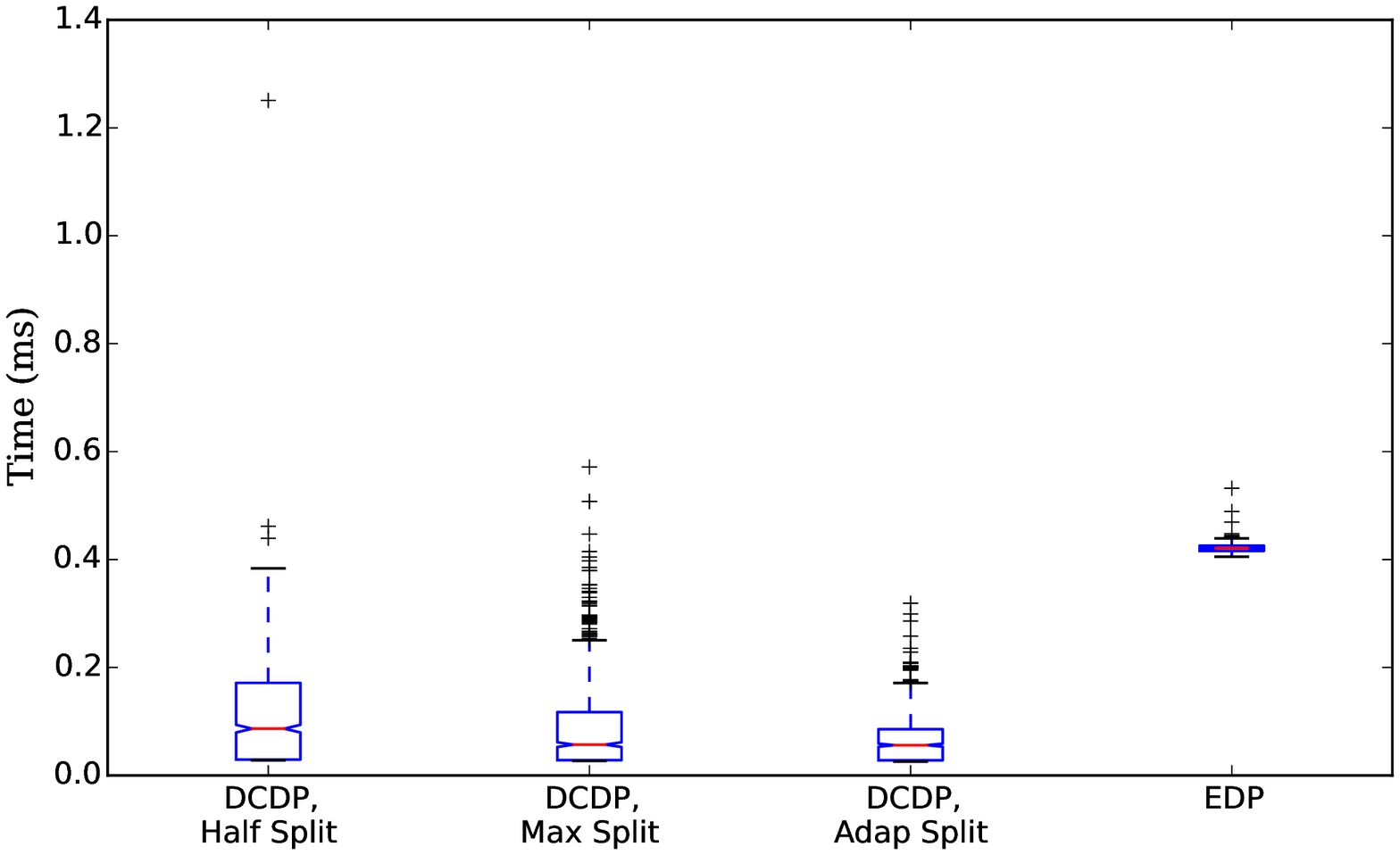}
\end{tabular}
\\
\end{center}
\caption{\textbf{Runtime Comparisons.}
In Panel(a), $n_{\text{dp}}$ of DCDP with three splitting schemes 
and EDP is shown, where $n_{\text{dp}}$ is the number of invoking the 
$O(mn\varsigma)$ DP routine. The computational time of each algorithm 
is plotted in Panel(b). 
\label{fig:demo86-03}}
\end{figure}

The computational time of the new segmentation algorithm,
DCDP, is compared with that of EDP. Note that the two
algorithms solve the same optimization problem, and it
can be shown that both algorithms always find an exact
optimal solution. DCDP and EDP are implemented in C++
language, and the runtimes are measured on a Linux machine
with Intel(R) Core(TM) i7 CPU and 8Gb memory. 
First, the number of times when the $O(nm\varsigma)$ DP
routine was invoked, which we denote by $n_{\text{dp}}$, is
counted using the annotated glomeruli in Set B.
Figure~\ref{fig:demo86-03}a shows the box-plot of
$n_{\text{dp}}$ for all methods. While the value of
$n_{\text{dp}}$ for EDP is always $n$, the values for DCDP
depend on the input images and the splitting schemes, Half
Split, Max Split, and Adap Split (
Section~\ref{ss:splitting-schemes}). For 46.32\% of
glomeruli, the optimal solutions are found within the first
DP routine (i.e. $n_{\text{dp}}=1$). The medians of the
$n_{\text{dp}}$\text{'}s when using Half Split, 
Max Split, and Adap Split are $5$, $3$, $3$, respectively. In other
words, the medians of the depths of the branching tree for
each scheme, respectively, are 3, 2 and 2. The 75 percentiles
of $n_{\text{dp}}$\text{'}s are $11$, $7$, $5$, respectively. 
For no glomeruli glomeruli, 
$n_{\text{dp}}$ of Adap Split is larger than $n$,  
whereas the number of glomeruli with
$n_{\text{dp}}>n$ are 4 (0.40\%) and 16 (1.61\%) for 
Half Split and Max Split. 
This implies that Adap Split is the smartest heuristic 
among the three splitting schemes. 
As considered in Subsection~\ref{ss:why-adap-split}, 
Adap Split produces the same
solution $\vp_{\text{L}}$ in the branches less frequently
than the three other schemes. In Half Split and Max Split,
the frequencies (\# of glomeluli) for the cases that the
solution $\vp_{\text{L}}$ in the top branch appears again
in the second branches is $414$ and $314$, respectively. 
Those numbers 
are much larger than the frequency in Adap Split which is only $97$.
This explains why Adap Split is faster. 
The actual runtimes of
each methods are depicted in Figure~\ref{fig:demo86-03}b,
where the medians of the computational times are 
0.0866, 0.0570, 0.0560, and 0.418 msec, respectively, 
for Half Split, Max Split, Adap Split and EDP. 
The 75 percentiles of the computational times are 
0.171, 0.117, 0.0856, and 0.426 msec, respectively. 
Since these values are
proportional to the $n_{\text{dp}}$\text{'}s, then the
ratios among the runtimes are almost same as the ratios
among the $n_{\text{dp}}$\text{'}s. These results conclude
that the proposed algorithm DCDP achieves an exact optimal
solution much more efficiently than the existing algorithm
EDP solves the same problem, and Adap Scheme is the fastest
splitting scheme.


\section{Conclusions}\label{s:conc}
In this paper, a new descriptor, Segmental HOG, was proposed for specific organ detection in microscopy images. The descriptor was based on the boundary of glomeruli to acquire robustness to variations of intensities, sizes, and shapes. A new segmentation algorithm, DCDP, was developed to locate the boundary of possible candidates of glomeruli. Empirical results show significant improvement compared to the state-of-the-art descriptor, Rectangular HOG, for the task of glomerulus detection in microscopy images. Moreover, experimental results reveal that DCDP is much faster than the existing segmentation algorithm EDP. 

Several possible extensions of the proposed method can be
considered. For instance, appropriate size of the sliding window
should be chosen if the proposed method is applied to
microscopic images with different resolutions. Also,
while the boundary likeliness function is
the same for any direction in the
segmentation algorithm, different boundary
likeliness functions can be used for detecting other organs
that have orientation. As for the block division of the
S-HOG descriptor, 24 blocks are used in this study as
depicted in Figure~\ref{fig:demo61-05-twohogs}c, but a
different number of blocks with a different division can be
used for another application. Future work includes exploring
such extensions in other applications of Segmental HOG.

\setcounter{algorithm}{0}
\setcounter{section}{0}
\renewcommand{\thealgorithm}{\Alph{algorithm}}
\renewcommand{\thesection}{\Alph{section}}


\section{Pruning}\label{ss:pruning}
Pruning can accelerate the DCDP algorithm. Consider the case where the lower bound $\ell$, such that
\begin{align*}
\max_{\vp\in\cS(\bN_{n})}J(\vp) \ge \ell, 
\end{align*}
is known in advance when searching for the solution in $\cS(\cI_{0})$. 
If $\ell > \max_{\vp\in\cS_{\text{L}}(\cI_{0})}J(\vp)$, 
then no optimal solution is in $\cS(\cI_{0})$ since 
\begin{align*}
\max_{\vp\in\cS(\bN_{n})}J(\vp) \ge \ell
> \max_{\vp\in\cS_{\text{L}}(\cI_{0})}J(\vp)
\ge \max_{\vp\in\cS(\cI_{0})}J(\vp). 
\end{align*}
Based on this fact, the pruning step is added to 
obtain Algorithm~\ref{algo:circopt2}, and we have the following lemma. 

\begin{lemma}\label{lem:opt-dcdp}
For any subset $\cI_{0}\subseteq\bN_{n}$ and 
$\forall \ell\in\bR\cup\{-\infty\}$, 
when the algorithm runs with
$(\vp_{0},J_{0},\ell_{0})=\textsc{DCDP}(\cI_{0},\ell)$, 
the returned tuple $(\vp_{0},J_{0},\ell_{0})$ satisfies one of 
the following:
\begin{itemize}
\item \textbf{Case G:} If $\max_{\vp\in\cS(\cI_{0})}J(\vp)\ge\ell$, then
\begin{align*}
\vp_{0}\in\argmax_{\vp\in\cS(\cI_{0})}J(\vp), \quad
\text{and }\quad
J_{0} = \ell_{0} = J(\vp_{0}) \ge \ell. 
\end{align*}
\item \textbf{Case L:} If $\max_{\vp\in\cS(\cI_{0})}J(\vp)<\ell$, then $J_{0} < \ell_{0} = \ell$.
\end{itemize}
\end{lemma}

\begin{algorithm}[th!]
\caption{
$(\vp_{0},J_{0},\ell_{0})=\textsc{DCDP}(\cI_{0},\ell)$. 
Modification of \textsc{DCDP\_Basic} by adding pruning steps. 
\label{algo:circopt2}}
\begin{algorithmic}[1]
{\normalsize
\REQUIRE $\cI_{0}\subseteq\bN_{n}$. 
\STATE $\vp_{\text{L}}\in\argmax_{\vp\in\cS_{\text{L}}(\cI_{0})}J(\vp)$; 
\IF{$\ell > J(\vp_{\text{L}})$} 
\STATE \COMMENT{Rule A}
\STATE $J_{0} := -\infty$; $\ell_{0} := \ell$; 
\STATE \textbf{return}; 
\ENDIF
\IF{$\vp_{\text{L}}\in\cS(\cI_{0})$}
\STATE \COMMENT{Rule B}
\STATE $\vp_{0} := \vp_{\text{L}}$; 
$J_{0} := J(\vp_{\text{L}})$; 
$\ell_{0} := \max( \ell, J_{0}) )$; 
\STATE \textbf{return}; 
\ENDIF
\STATE \COMMENT{Rule C}
\STATE $(\cI_{1},\cI_{2}) := \textsc{Split}(\cI_{0})$; 
\STATE $(\vp_{1},J_{1},\ell_{1}) := \textsc{DCDP}(\cI_{1},\ell)$;  
\STATE $(\vp_{2},J_{2},\ell_{2}) := \textsc{DCDP}(\cI_{2},\ell_{1})$;  
\STATE $i^{\star} \in \argmax_{i\in\{1,2\}}J_{i}$; 
\STATE $\vp_{0} := \vp_{i^{\star}}$; $J_{0} := J(\vp_{i^{\star}})$; 
\STATE $\ell_{0} := \max( \ell_{2}, J_{0} )$; 
}
\end{algorithmic}
\end{algorithm}

Section~\ref{ss:proof-of-lem-opt-dcdp} gives the proof of the lemma. Finally, from Lemma~\ref{lem:opt-dcdp}, we can derive the following theorem which is an important theoretical result of this study. 

\begin{theorem}
The optimal solution of the problem \eqref{eq:origprob-glomdp} is 
obtained by invoking 
$(\vp^{\star},J^{\star},\ell^{\star})=\textsc{DCDP}(\bN_{n},-\infty)$. 
\end{theorem}

\section{Splitting Schemes}\label{ss:splitting-schemes}
For the implementation of $\textsc{Split}(\cI_{0})$, 
we considered three schemes: 
\textit{Half Split}, 
\textit{Max Split}, and
\textit{Adap Split}.

\paragraph{Half Split.}  In this scheme, the subset of indices $\cI_{0}$
is simply divided into the first half and the second half. 
The resulting $\cI_{1}$ and $\cI_{2}$ are 
sets of consecutive integers.  For instance, this scheme divides $\cI_{0} = \{ 7,8,9,10,11,12 \}$ into $\cI_{1} = \{ 7,8,9 \}$ and $\cI_{2} = \{ 10,11,12 \}$. 
To increase the lower-bound $\ell$ defined earlier, a heuristic that swaps $\cI_{1}$ with $\cI_{2}$ if 
\begin{align*}
\max_{h\in\cI_{1}}L_{m}(h) < \max_{h\in\cI_{2}}L_{m}(h)  
\end{align*}
is employed. 


\paragraph{Max Split.}  Similar to Half Split, 
$\cI_{1}$ and $\cI_{2}$ generated by Max Split are sets of 
consecutive integers, although the splitting points are 
different.  The splitting point of Half Split is 
the center of the interval~$\cI_{0}$, whereas 
the splitting point of Max Split is given by
$h_{\star}(\cI_{0}) := \argmax_{h\in\cI_{0}}L_{m}(h)$.  
For example, 
if $h_{\star}(\cI_{0})=8$ 
for $\cI_{0} = \{ 7,8,9,10,11,12 \}$, this scheme outputs $\cI_{1} = \{ 7,8 \}$ and $\cI_{2} = \{ 9,10,11,12 \}$. 
In general, the resulting divisions are given by
\begin{align*}
\cI_{1} := \left\{ h\in\cI_{0}\,|\, h\le h_{\star}(\cI_{0}) \right\},
\quad\text{ and }\quad
\cI_{2} := \left\{ h\in\cI_{0}\,|\, h> h_{\star}(\cI_{0}) \right\}. 
\end{align*}
If $\text{card}(\cI_{2})=0$, the entry 
$h_{\star}(\cI_{0})$ is moved from $\cI_{1}$ to $\cI_{2}$. 
A heuristic that swaps $\cI_{1}$ with $\cI_{2}$ if 
$\text{card}(\cI_{1}) >  \text{card}(\cI_{2})$
is applied. 

\paragraph{Adap Split.}  
Different from the above two schemes, Adap Scheme adaptively 
determines the splitting point using the current solution 
$\vp_{\text{L,0}}:=\argmax_{\vp\in\cS_{\text{L}(\cI_{0})}}J(\vp)$. 
Let us denote 
the first entry and the last entry in the vector $\vp_{\text{L,0}}$
by $p^{0}_{1}$ and $p^{0}_{m}$, respectively. The splitting 
point of Adap Scheme is given by the center between $p^{0}_{1}$ 
and $p^{0}_{m}$. For instance, when $p^{0}_{1}=9$ and 
$p^{0}_{m}=12$ $\cI_{0}=\{7,8,9,10,11,12\}$, this splitting 
scheme divides $\cI_{0}$ into $\cI_{1}=\{7,8,9,10\}$ and 
$\cI_{2}=\{11,12 \}$. 
In general, the resulting divisions are given by
\begin{align*}
\cI_{1} :=  
\left\{ h\in\cI_{0}\,|\, 
h\le \frac{p^{0}_{1}+p^{0}_{m}}{2} \right\},
\quad\text{ and }\quad
\cI_{2} :=  
\left\{ h\in\cI_{0}\,|\, 
h> \frac{p^{0}_{1}+p^{0}_{m}}{2} \right\}. 
\end{align*}
The smallest entry in $\cI_{2}$ is moved to $\cI_{1}$ 
if $\text{card}(\cI_{1})=0$. 
The largest entry in $\cI_{1}$ is moved to $\cI_{2}$ 
if $\text{card}(\cI_{2})=0$. 
A swapping heuristic used in Max Split is then applied. 

\subsection{Why Adap Split is better}  
\label{ss:why-adap-split}
\begin{figure}[ht]
\begin{center}
\begin{tabular}{ll}
(a) $\cS(\bN_{8})$ & (b) $\cS_{\text{L}}(\bN_{8})$ 
\\
\includegraphics[width=0.30\textwidth]{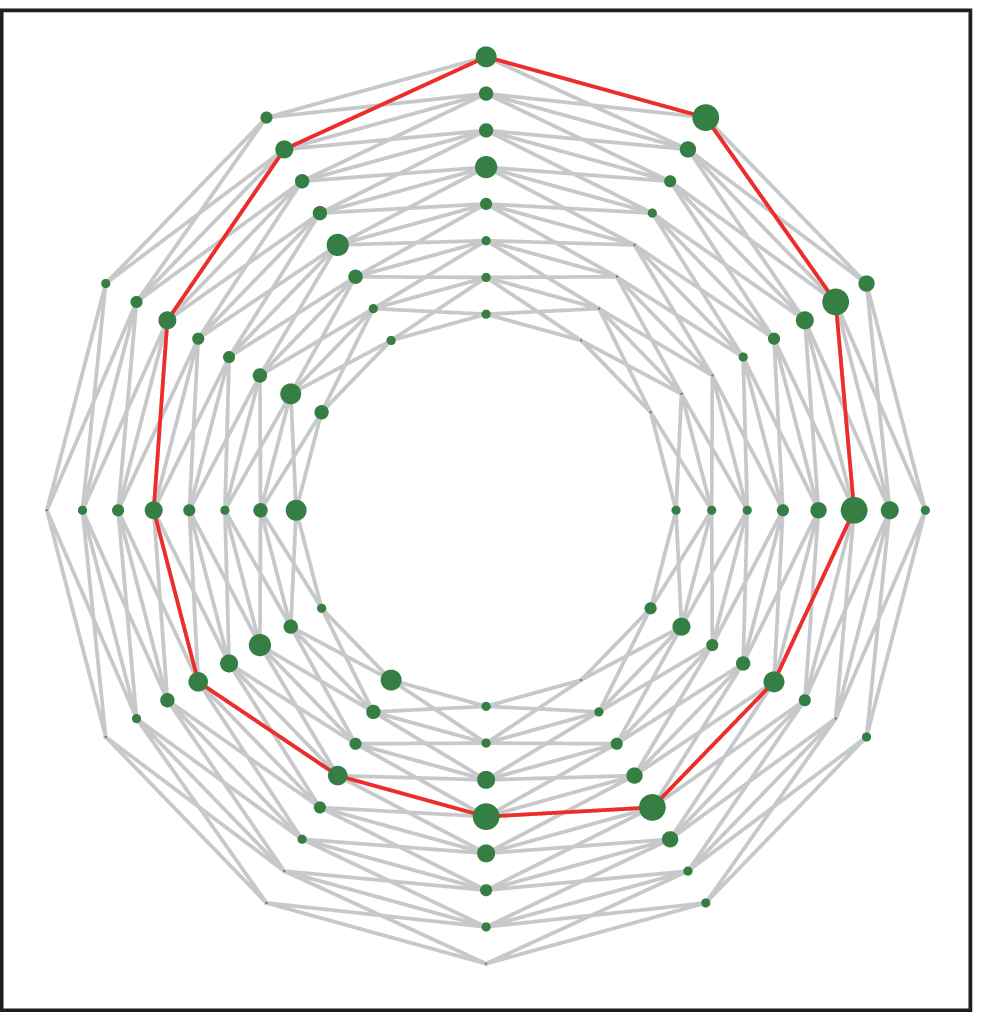}
&
\includegraphics[width=0.30\textwidth]{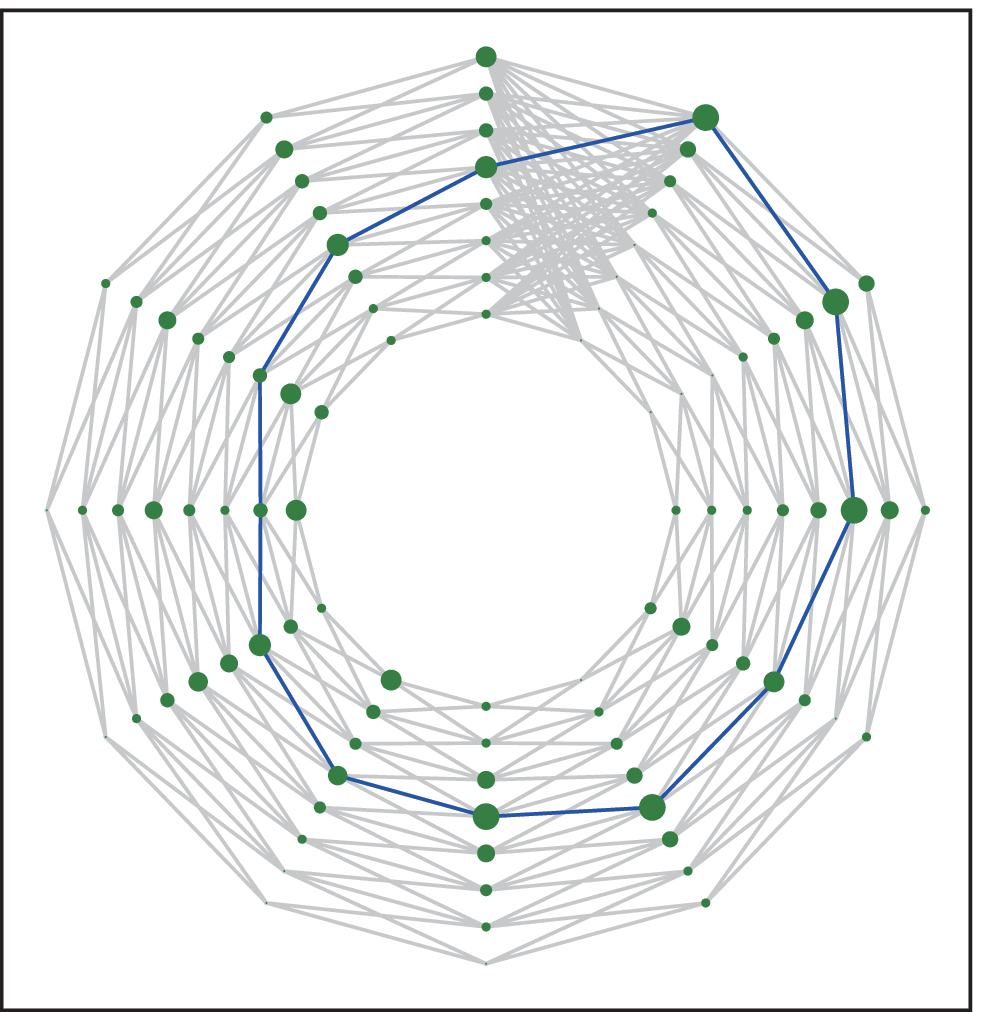}
\\
\\
(c) $\cS_{\text{L}}(\{7,8\})$  & (d) $\cS_{\text{L}}(\{1,2,\dots,6\})$ 
\\
\includegraphics[width=0.30\textwidth]{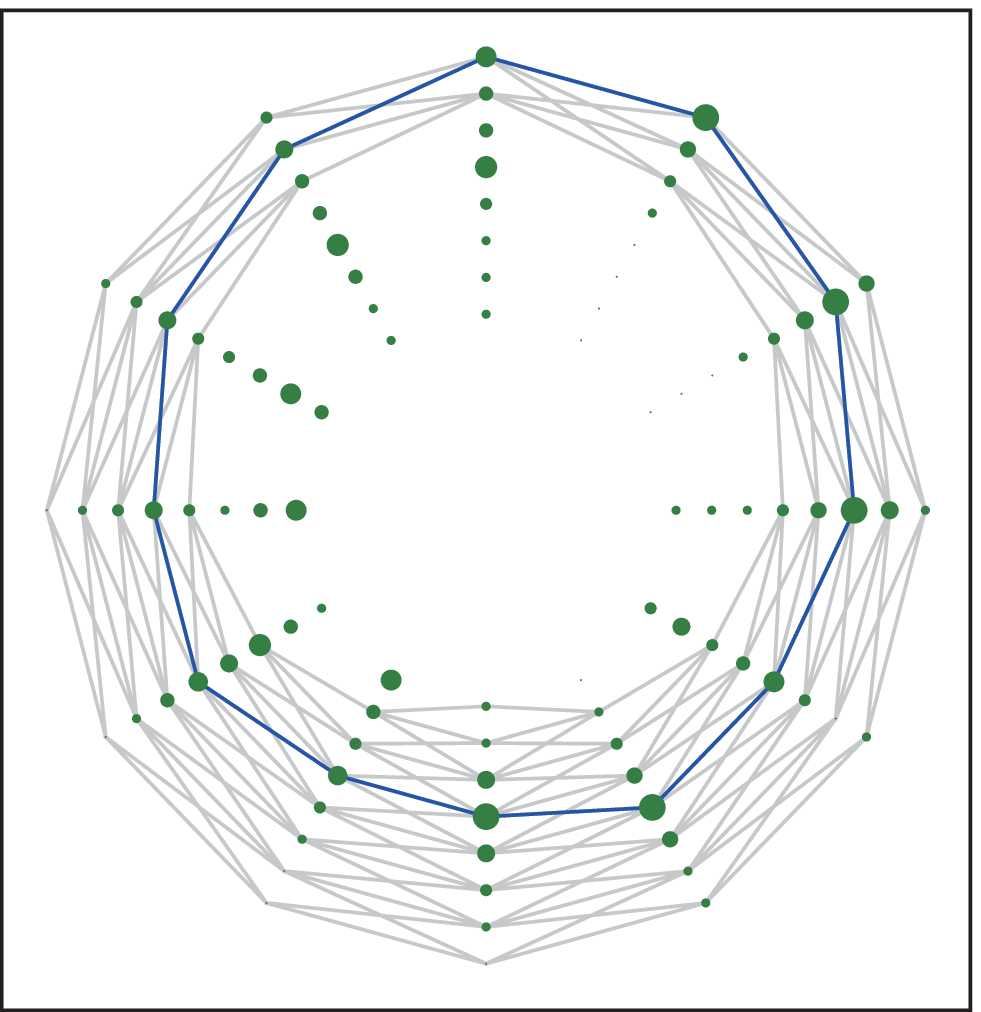}
&
\includegraphics[width=0.30\textwidth]{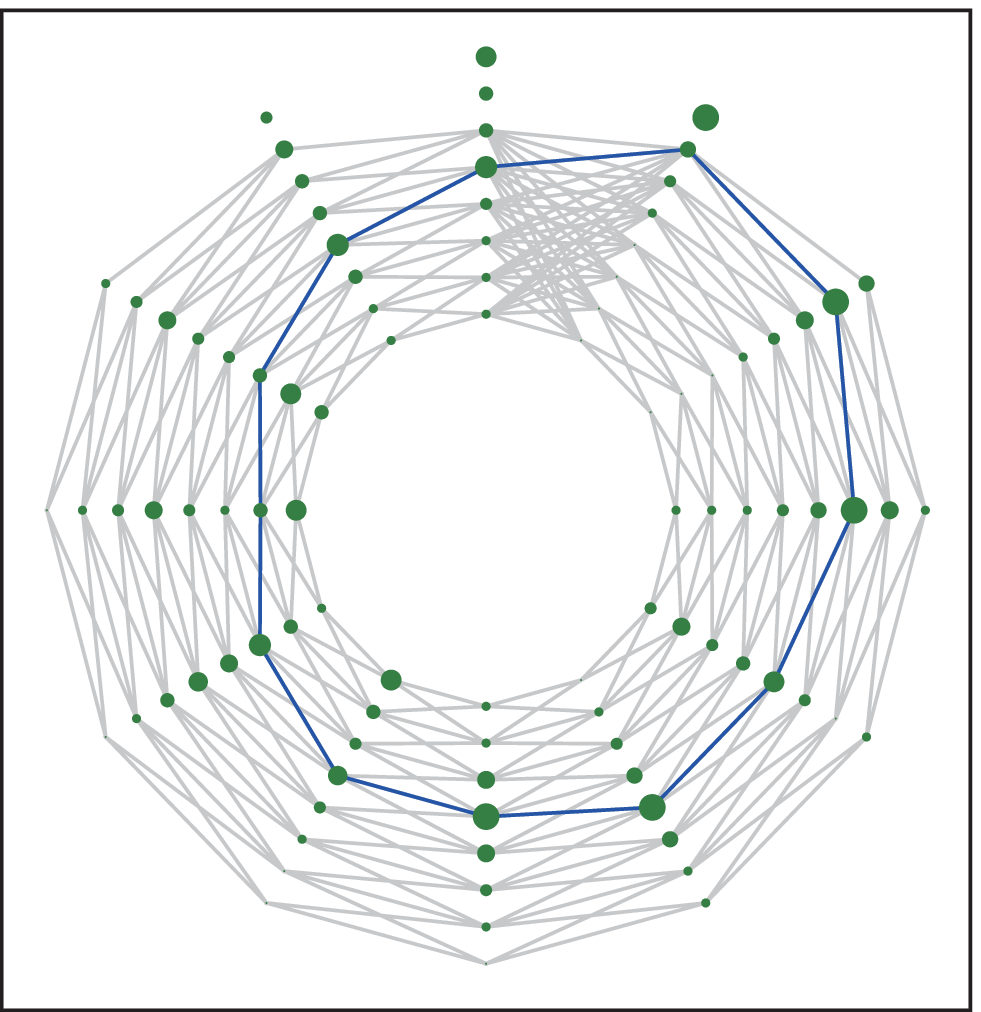}
\\
\\
(e) $\cS_{\text{L}}(\{5,6,7,8\})$  & (f) $\cS_{\text{L}}(\{1,2,3,4\})$ 
\\
\includegraphics[width=0.30\textwidth]{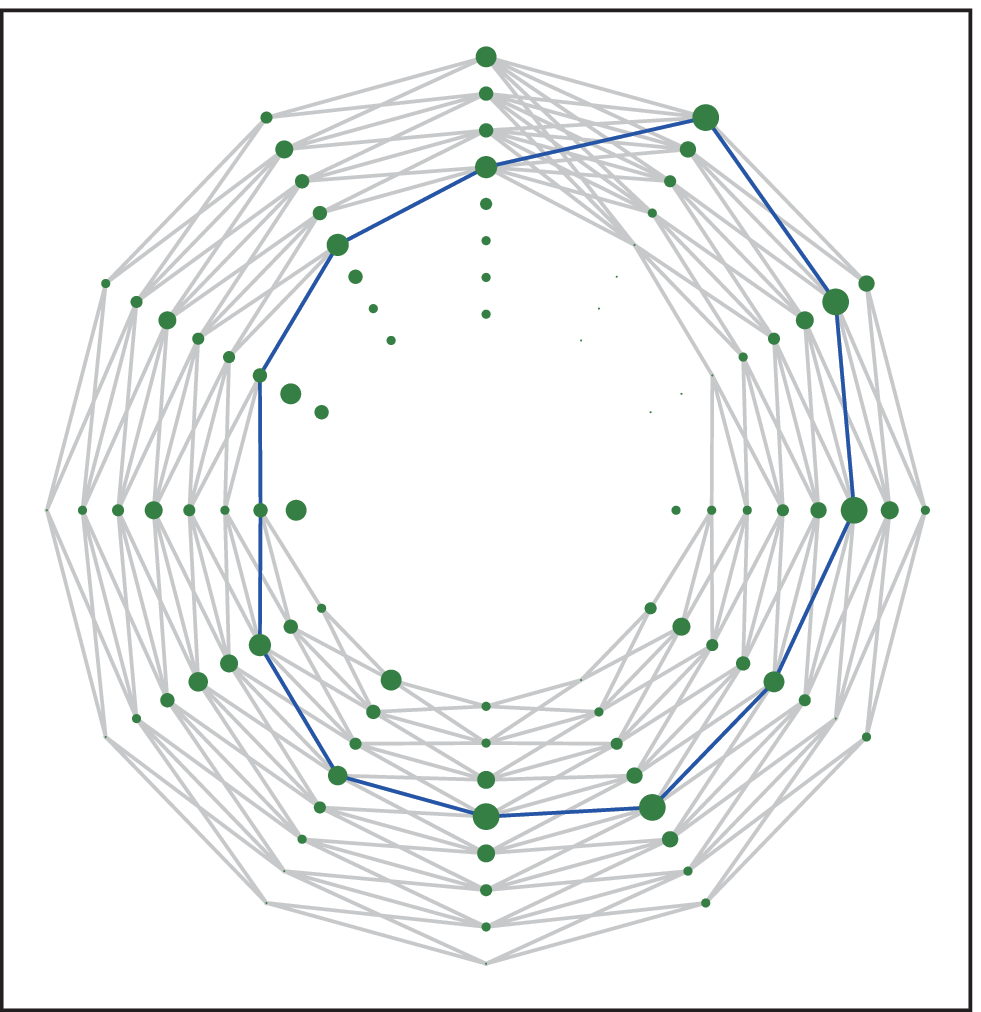}
&
\includegraphics[width=0.30\textwidth]{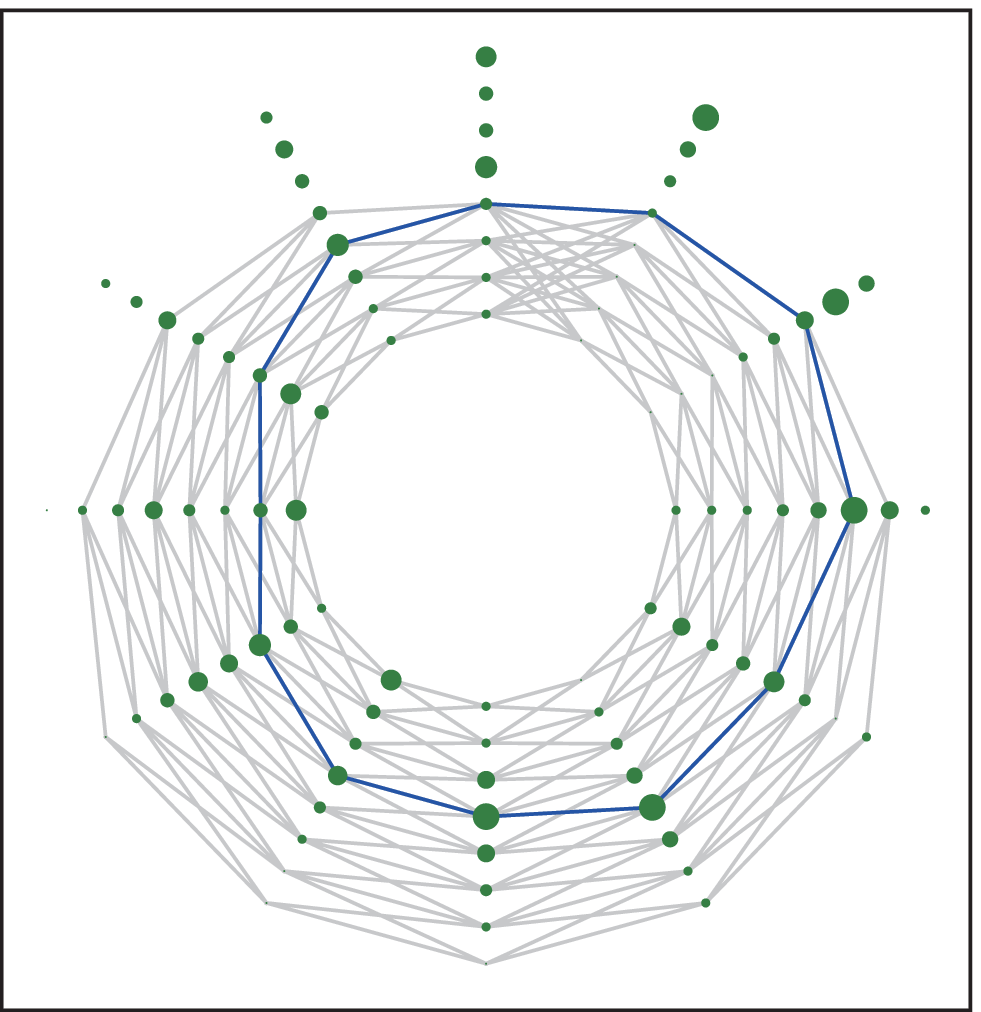}
\\
\end{tabular}
\end{center}
\caption{\textbf{Relaxed Problems in DCDP.}
Here a segmentation problem~\eqref{eq:origprob-glomdp} 
with $(n,m,\varsigma)=(12,8,1)$ 
is considered. In Panel (a), the sizes of red circles indicate the 
quantities of boundary likeliness. The feasible region $\cS(\cI_{0})$ 
is the set of polygons, where $\cI_{0}=\bN_{n}$. Overlapping all the 
polygons yields the gray edges. The optimal polygon is drawn with 
the red edges. DCDP relaxes the feasible region to get 
$\cS_{\text{L}}(\cI_{0})$. In Panel (b), the relaxed feasible region 
$\cS_{\text{L}}(\cI_{0})$ is depicted. The blue polygon is the optimal 
solution for the relaxed problem 
$\vp_{\text{L},0}=\argmax_{\vp\in\cS_{\text{L}}(\cI_{0})}J(\vp)$. 
The relaxed problems of four sub-problems 
with $\cS_{\text{L}}(\{7,8\})$, $\cS_{\text{L}}(\{1,2,\dots,6\})$, 
$\cS_{\text{L}}(\{5,6,7,8\})$, and $\cS_{\text{L}}(\{1,2,3,4\})$ 
are illustrated in 
Panels (c),(d),(e),(f). 
The blue polygons in (c),(d),(e),(f) are the optimal solution 
of the four relaxed problems, respectively. 
\label{fig:demo213-03}}
\end{figure}

Adap Scheme is expected to be the smartest heuristic 
among the three splitting schemes. To describe the reason, 
let us illustrate the process of DCDP on a small toy problem 
with $(n,m,\varsigma)=(12,8,1)$ shown in Figure~\ref{fig:demo213-03}. 
The original problem and the relaxed problem are depicted in 
Figure~\ref{fig:demo213-03}a,b, respectively. 
When running $\textsc{DCDP}(\bN_{8},-\infty)$, 
it is observed that 
$\vp_{\text{L},0}:=\argmax_{\vp\in\cS_{\text{L}}(\cI_{0})}J(\vp)\not\in\cS(\cI_{0})$,  
and thereby the set $\cI_{0}$ is divided into $\cI_{1}$ and $\cI_{2}$
to produce two new branches 
$\textsc{DCDP}(\cI_{1},-\infty)$ and $\textsc{DCDP}(\cI_{2},\ell_{1})$
where $\ell_{1}$ will be computed by the former branch 
$\textsc{DCDP}(\cI_{1},-\infty)$.  

If using the Adap Split scheme, the two subsets are
$\cI_{1}=\{7,8\}$ and $\cI_{2}=\{1,2,3,4,5,6\}$. 
In the branch of $\textsc{DCDP}(\cI_{1},-\infty)$, 
$\vp_{\text{L},1}:=\argmax_{\vp\in\cS_{\text{L}}(\cI_{1})}J(\vp)$ 
is in $\cS (\cI_{1})$ (as shown in Figure~\ref{fig:demo213-03}c), 
implying that 
$\vp_{\text{L},1}$ is the maximizer of $J(\vp)$ over $\cS(\cI_{1})$
and no more branching occurs. $\ell_{1}=J(\vp_{\text{L},1})=10.1$
is set. 
Next, $\textsc{DCDP}(\cI_{2},\ell_{1})$ is invoked and 
$\vp_{\text{L},2}:=\argmax_{\vp\in\cS_{\text{L}}(\cI_{2})}J(\vp)$ 
is computed in the branch (Figure~\ref{fig:demo213-03}d). 
It is then observed that $J(\vp_{\text{L},2})=9.5 < 10.1 = \ell_{1}$, 
which implies that the optimal solution is not in $\cS(\cI_{2})$. 
Hence, $\vp_{\text{L},1}$ turns out the optimal solution of 
the original solution. 

Meanwhile, if Half Split is applied, the set
$\cI_{0}=\bN_{8}$ is divided into $\cI_{1}=\{5,6,7,8\}$ and 
$\cI_{2}=\{1,2,3,4\}$ (Figure~\ref{fig:demo213-03}e,f). 
In the branch of $\textsc{DCDP}(\cI_{1},-\infty)$, the solution 
of the relaxed problem is again 
$\vp_{\text{L},1}:=\argmax_{\vp\in\cS_{\text{L}}(\cI_{1})}J(\vp)=\vp_{\text{L},0}\not\in\cS(\cI_{1})$, 
leading to further branching of this branch (Figure~\ref{fig:demo213-03}e). 
Actually, in our experiments described in Results section, 
it is observed that Half Split and Max Split frequently 
encounter $\vp_{\text{L},1}=\vp_{\text{L},0}$ or 
$\vp_{\text{L},2}=\vp_{\text{L},0}$. Whenever $\vp_{\text{L},1}=\vp_{\text{L},0}$, 
further new branches for the divisions of $\cI_{1}$ are produced, 
because $\vp_{\text{L},1}=\vp_{\text{L},0}\not\in\cS(\cI_{0})$ 
leading to $\vp_{\text{L},1}\not\in\cS(\cI_{1})\subset\cS(\cI_{0})$. 
Similarly, when $\vp_{\text{L},2}=\vp_{\text{L},0}$, 
further new branches for the divisions of $\cI_{1}$ are always born. 

In Adap Split, as desribed in Results section, 
$\vp_{\text{L},1}=\vp_{\text{L},0}$ or 
$\vp_{\text{L},2}=\vp_{\text{L},0}$ is 
less likely to happen, resulting in less branches.

\section{Proof of Lemma~\ref{lem:opt-dcdp}}
\label{ss:proof-of-lem-opt-dcdp}
We shall use the following notation: For any $\cI\subseteq\bN_{n}$, 
\begin{align*}
J(\cI) := \max_{\vp\in\cS(\cI)}J(\vp),
\quad\text{ and }\quad
J_{\text{L}}(\cI) := \max_{\vp\in\cS_{\text{L}}(\cI)}J(\vp). 
\end{align*}
The following relationships will be used in this proof: 
\begin{align*}
J(\vp_{\text{L}}) \stackrel{\smash{\mathrm{(eqA)}}}{=} J_{\text{L}}(\cI_{0}) 
\stackrel{\smash{\mathrm{(ineqA)}}}{\ge} J(\cI_{0}) \stackrel{\smash{\mathrm{(eqB)}}}{=} \max(J(\cI_{1}),J(\cI_{2})), 
\end{align*}
where the labels (eqA), (ineqA), 
and (eqB) are used to distinguish these equalities and the inequality in later descriptions of this proof.
The inequality follows from the fact that 
$\cS(\cI_{0})\subseteq\cS_{\text{L}}(\cI_{0})$, 
while the second equality eqB follows from 
$\cS(\cI_{0}) = \cS(\cI_{1}) \cup \cS(\cI_{2})$. 

We will prove the lemma by induction. For the case where 
$\text{card}(\cI_{0})=1$, observe that  
$\cS_{\text{L}}(\cI_{0})=\cS(\cI_{0})$, implying 
that 
\begin{align*}
\vp_{\text{L}}\in\argmax_{\vp\in\cS(\cI_{0})}J(\vp)
\quad\text{ and }\quad
J(\cI_{0}) = J_{\text{L}}(\cI_{0}) = J_{0}.  
\end{align*}
%
If $J(\cI_{0}) = J(\vp_{\text{L}})<\ell$, then by Rule A in the algorithm, $J_{0}=-\infty$ and $J_{0} < \ell_{0}= \ell$.
On the other hand, if $J(\vp_{\text{L}})\ge\ell$, then since $\vp_{\text{L}}\in\argmax_{\vp\in\cS(\cI_{0})}J(\vp)$, we have
$\vp_{\text{L}}\in\cS(\cI_{0})$. Thus, by Rule B, we have $\vp_{0}=\vp_{\text{L}}$ and $J_{0}=\ell_{0}\ge\ell$.
Therefore, the lemma is true for $\text{card}(\cI_{0})=1$.

Let us now assume that the lemma holds
for any $\cI_{0}\subseteq \bN_{n}$ such that
$\text{card}(\cI_{0}) < k$
to show that the lemma is also established 
for any $\cI_{0}$ such that $\text{card}(\cI_{0})=k$.
Now suppose that  $\cI_{0}\subseteq \bN_{n}$ and $\text{card}(\cI_{0}) = k$.
The following is an exhaustive list of all possible cases: 
\begin{enumerate}
\item[(1)]
$\ell > J_{\text{L}}(\cI_{0})$, 
\item[(2)] 
$J_{\text{L}}(\cI_{0}) \ge \ell$ and $\vp_{\text{L}}\in\cS(\cI_{0})$,  
\item[(3)] 
$J_{\text{L}}(\cI_{0}) \ge J(\cI_{0}) = J(\cI_{1}) = J(\cI_{2}) \ge \ell$
and 
$\vp_{\text{L}}\not\in\cS(\cI_{0})$, 
\item[(4)] 
$J_{\text{L}}(\cI_{0})\ge J(\cI_{0}) = J(\cI_{1}) \ge \ell > J(\cI_{2})$ and 
$\vp_{\text{L}}\not\in\cS(\cI_{0})$, 
\item[(5)]
$J_{\text{L}}(\cI_{0})\ge J(\cI_{0}) = J(\cI_{1}) > J(\cI_{2}) \ge \ell $ and 
$\vp_{\text{L}}\not\in\cS(\cI_{0})$, 
\item[(6)] 
$J_{\text{L}}(\cI_{0}) \ge J(\cI_{0}) = J(\cI_{2}) > J(\cI_{1}) \ge \ell$
and 
$\vp_{\text{L}}\not\in\cS(\cI_{0})$, 
\item[(7)] 
$J_{\text{L}}(\cI_{0}) \ge J(\cI_{0}) = J(\cI_{2}) \ge \ell> J(\cI_{1})$
and 
$\vp_{\text{L}}\not\in\cS(\cI_{0})$, 
\item[(8)] 
$J_{\text{L}}(\cI_{0}) \ge \ell > J(\cI_{0})$
and
$\vp_{\text{L}}\not\in\cS(\cI_{0})$. 
\end{enumerate}
We shall show that for each of the seven cases above, either Case G or Case L is true. 

\begin{enumerate}
%
\item[(1)] Since
$\ell > J_{\text{L}}(\cI_{0}) \stackrel{\smash{\mathrm{(eqA)}}}{=} J(\vp_{\text{L}}) \stackrel{\smash{\mathrm{(ineqA)}}}{\ge} J(\cI_{0})$, 
then by Rule A, we have $J_{0} = -\infty$ and $J_{0}<\ell_{0}=\ell$. Thus, Case L is satisfied. 

%
\item[(2)]
If $J_{\text{L}}(\cI_{0}) \ge \ell$, then $J_(\vp_{\text{L}}) \ge \ell$.
Moreover, $\vp_{\text{L}}\in\cS(\cI_{0})$ implies $\vp_{\text{L}}\in\argmax_{\vp\in\cS(\cI_{0})}J(\vp)$ 
and $J(\vp_{\text{L}}) = J(\cI_{0})$. 
Then by Rule B, 
$J_{0} = J(\vp_{\text{L}}) = J(\cI_{0}) = \ell_{0}$ 
and 
$\vp_{0} = \vp_{\text{L}}\in\argmax_{\vp\in\cS(\cI_{0})}J(\vp)$. Therefore, Case G holds. 

%
\item[(3)] 
Given $J(\vp_{\text{L}}) \stackrel{\smash{\mathrm{(eqA)}}}{=} J_{\text{L}}(\cI_{0})\ge \ell$ and $\vp_{\text{L}}\not\in\cS(\cI_{0})$,
Rule C is applied. Observe that $(\vp_{1},J_{1},\ell_{1})$ satisfies Case G since $J(\cI_{1})=\max_{\vp\in\cS(\cI_{1})}J(\vp)\ge \ell$.
Thus, $J_{1}=\ell_{1}=J(\cI_{1})\ge\ell$. Similarly, $(\vp_{2},J_{2},\ell_{2})$ also satisfies Case G, and 
we have $J_{2}=\ell_{2}=J(\cI_{2})\stackrel{\smash{\mathrm{(eqB)}}}{=}J(\cI_{1})=J(\cI_{0})$. 
Therefore, the returned value $\vp_{0}$ of $\mbox{DCDP}(\cI_{0},\ell)$ 
is equal to $\vp_{1}$ or to $\vp_{2}$, 
both of which are in $\argmax_{\vp\in\cS(\cI_{0})}J(\vp)$. Furthermore, we obtain 
$\ell_{0}=J_{0}=J_{1}=J_{2}=J(\cI_{0})\ge\ell$. Hence, we have Case G.

%
\item[(4)] 
As with Case (3), $(\vp_{1},J_{1},\ell_{1})$ satisfies Case G, and we have $\vp_{1}\in\argmax_{\vp\in\cS(\cI_{1})}J(\vp)$ and
$J_{1}=\ell_{1}=J(\cI_{1})\ge\ell$.
An so it follows that $\max_{\vp\in\cS(\cI_2)}J(\vp_{2})=J(\cI_2) < \ell \leq \ell_{1}$,  
and Case L holds for $(\vp_{2},J_{2},\ell_{2})$ with $J_{2}<\ell_{2}=\ell_{1}$. 
Therefore, output 
$\vp_{0} = \vp_{1}$, which is in $\argmax_{\vp\in\cS(\cI_{0})}J(\vp)$ since $\cS(\cI_{1})\subset\cS(\cI_{0})$, and
$J_{0}=J_{1}=\ell_{2}=\ell_{0}\ge \ell$. This gives us Case G. 

\item[(5)] 
In a similar logic as in Case (4), Case G is true for $(\vp_{1},J_{1},\ell_{1})$, and $\vp_{1}\in\argmax_{\vp\in\cS(\cI_{1})}J(\vp)$ and
$J_{1}=\ell_{1}=J(\cI_{1})\ge\ell$. Given that $J(\cI_{2}) < J(\cI_{1})$, then $J(\cI_{2})<\ell_{1}$, hence, Case L also holds for $(\vp_{2},J_{2},\ell_{2})$
in this sub-case. Therefore, the same conclusion from Case (4) follows.  

%
\item[(6)]
Since $\vp_{\text{L}}\not\in\cS(\cI_{0})$, Rule C is implemented.
Note that with $\max_{\vp\in\cS(\cI_{1})}J(\vp)=J(\cI_{1})\ge \ell$,
$(\vp_{1},J_{1},\ell_{1})$ follows Case G, and we have $J_{1}=\ell_{1}=J(\cI_{1})\ge\ell$.
This implies that $J(\cI_{2})>J(\cI_{1})=\ell_{1}$. Therefore, Case G also holds for
$(\vp_{2},J_{2},\ell_{2})$, and we obtain 
$J_{2}=\ell_{2}=J(\cI_{2})>J(\cI_{1})=J_{1}$. 
Thus, $\mbox{DCDP}(\cI_{0},\ell)$ outputs 
$\vp_{0} = \vp_{2} \in\argmax_{\vp\in\cS(\cI_{0})}J(\vp)$ and
$J_{0}=J_{2}=\ell_{2}=\ell_{0}>\ell$. Hence, Case G holds. 

%
\item[(7)]
Similar to the previous cases, we apply Rule C.
Given $\ell> J(\cI_{1})=\max_{\vp\in\cS(\cI_{1})}J(\vp)$, by Case L, $J_{1}<\ell_{1}=\ell$. 
and so it follows that $J(\cI_{2})\ge \ell_{1}=\ell$. Therefore, $(\vp_{2},J_{2},\ell_{2})$ satisfies Case G, 
with $J_{2}=\ell_{2}=J(\cI_{2})>\ell$. 
Hence, the returned values of $\mbox{DCDP}(\cI_{0},\ell)$ are 
$\vp_{0} = \vp_{2} \in\argmax_{\vp\in\cS(\cI_{0})}J(\vp)$ and
$J_{0}=J_{2}=\ell_{2}=\ell_{0}>\ell$, and Case G is satisfied. 

%
\item[(8)]
Likewise, we implement Rule C for this case. 
Observe that since $\max(J(\cI_{1}),J(\cI_{2})) \stackrel{\smash{\mathrm{(eqB)}}}{=} J(\cI_{0}) <\ell$, then both 
$(\vp_{1},J_{1},\ell_{1})$ and $(\vp_{2},J_{2},\ell_{2})$ satisfy Case L. Thus, we have
$J_{1}<\ell_{1}=\ell$ and $J_{2}<\ell_{1}=\ell_{2}$. 
Therefore, $\mbox{DCDP}(\cI_{0},\ell)$ outputs 
$J_{0}=\max(J_{1},J_{2})<\ell_{2}=\ell_{0}=\ell_{1}=\ell$, which corresponds to Case L.

\end{enumerate}


\bibliographystyle{plain}

\newpage

\end{document}